%% file: hs2.tex
\documentclass[12pt]{article}

\input preamble.tex

\title{An Empirical Analysis of Federated Learning Models Subject to Label-Flipping Adversarial Attack}

\author{Kunal Bhatnagar\footnotemark[1]\ \ \  
Sagana Chattanathan\footnotemark[1]\ \ \ 
Angela Dang\footnotemark[1]\ \ \ \\[0.75ex]
Bhargav Eranki\footnotemark[1]\ \ \ 
Ronnit Rana\footnotemark[1]\ \ \ 
Charan Sridhar\footnotemark[1]\ \ \ \\[0.75ex]
Siddharth Vedam\footnotemark[1]\ \ \ 
Angie Yao\footnotemark[1]\ \ \ 
Mark Stamp\footnotemark[1]\,\,\footnotemark[2]}

\begin{document}

\symbolfootnotetext[1]{Department of Computer Science, San Jose State University}
\symbolfootnotetext[2]{mark.stamp$@$sjsu.edu}

\maketitle

\abstract
In this paper, we empirically analyze adversarial attacks on selected federated learning 
models. The specific learning models considered are
Multinominal Logistic Regression (MLR),
Support Vector Classifier (SVC),
Multilayer Perceptron (MLP),
Convolution Neural Network (CNN),
Random Forest,
XGBoost,
and 
Long Short-Term Memory (LSTM).
For each model, we simulate label-flipping attacks,
experimenting extensively with~10 federated clients
and~100 federated clients. We vary the percentage of adversarial clients
from~10\%\ to~100\%\ and, simultaneously, the percentage of labels
flipped by each adversarial client is also varied from~10\%\ to~100\%.
Among other results, we find that models differ in their inherent
robustness to the two vectors in our label-flipping attack, i.e., the percentage of 
adversarial clients, and the percentage of labels flipped by each adversarial client. 
We discuss the potential practical implications of our results.

\section{Introduction}\label{chap:Introduction}

The Federated Learning (FL) paradigm offers the advantage of maintaining the privacy of local 
training data, while also distributing some of the work required to train models. Although the 
accuracy of FL models tends to be lower than models trained via traditional centralized 
learning techniques, the tradeoff may be worthwhile in many cases, especially in situations
where data privacy would otherwise make training models impractical.

The distributed nature of FL opens the door to a wide range of adversarial attack scenarios.
In this paper, we empirically analyze the effectiveness of label-flipping attacks.
We simulate such attacks by assigning a percentage 
of clients as adversarial, with each adversarial client
flipping a specified percentage of the labels in its local training dataset.
We experiment with seven distinct FL models, namely,
Multinominal Logistic Regression (MLR),
Support Vector Classifier (SVC),
Multilayer Perceptron (MLP),
Convolution Neural Network (CNN),
Random Forest,
XGBoost,
and
Long Short-Term Memory (LSTM).
For each of these models, we carefully analyze the case with~10 federated clients 
and the case with~100 federated clients, giving us~14 experiments, in total. 
Furthermore, for each of these~14 experiments,
the percentage of adversarial clients ranges from~10\%\ to~100\%,
while the label flipping percentage simultaneously ranges from~10\%\ to~100\%,
giving us~100 data points per experiment. 

For each experiment, we provide a 3-dimensional graph of the accuracy 
as a function of both the percentage of adversarial clients and the percentage
of labels flipped. 
We further analyze our results and show that
some models are inherently more robust with respect to the percentage 
of adversarial clients, while other models are more robust with respect to the percentage of
labels flipped by each adversarial client.
That is, for a given overall percentage of labels flipped,
some models retain more of their accuracy when relatively few adversarial
clients flip relatively many labels, whereas other models
retain more of their accuracy when the converse is true.
This has potential practical implications, as we might choose to favor 
specific federated learning models for a given application based on 
likely attack scenarios or available defensive techniques.

The remainder of this paper is organized as follows. In Section~\ref{chap:2}, we provide
some background information on FL, and we introduce the specific learning models
considered in this paper. Section~\ref{chap:3} provides implementation details, with
the emphasis on the dataset used in our experiments and our experimental design.
Our experimental results are presented and discussed in Section~\ref{chap:4}.
Section~\ref{chap:5} concludes the paper and considers potential directions for future work.

\section{Background}\label{chap:2}

In this section, we present a brief introduction to relevant aspects of Federated Learning (FL).
Among other topics, we discuss the aggregation strategy used 
in our FL experiments. We also introduce the specific FL models that are 
considered in this paper. 

\subsection{Federated Learning}

Federated learning (FL) models are trained in a distributed manner, where  
the data is decentralized among a number of clients. The clients train local
models on their local data and, typically, a central server periodically collects these model parameters 
(e.g., weights). The central server then aggregates the parameters to build an overall model. This is  
in contrast to a traditional machine learning environment, where all data and computing resources are 
centralized. 

FL is not to be confused with distributed learning. In distributed learning, training is parallelized
across multiple servers, and the dataset at each client is assumed to be identically distributed 
and approximately the same size. In FL, the dataset at clients can be heterogeneous in terms of size
and other aspects, e.g., only a subset of classes might be present in a given client's 
dataset~\cite{pmlr-v54-mcmahan17a}. 

In this paper, we consider an FL training process consisting of multiple rounds 
coordinated by a centralized server. Each round consists of the following steps.
\begin{enumerate}
\item {\bf Broadcast}: The clients download the current ML model and global weights from the server.
\item {\bf Client Computation}: Each client instantiates the training model using the downloaded weights and 
conducts local training on their local dataset. 
\item {\bf Aggregation}: The client model updates are aggregated by the server using an aggregating strategy. 
\item {\bf Model Update}: The aggregated weights are used to update the global model and the global 
model is evaluated to determine if this round has produced an improved model. 
\end{enumerate}
Note that multiple rounds are needed, as the global model updates 
are computed on the centralized server.

The aggregating strategy is a key component of FL training process outlined above. 
In this paper, we use a federated averaging (FedAvg) approach. As the name suggests,
FedAvg involves computing the average of the client model weights. 
The intuition is that averaging the model weights has a similar effect of the model gradients.

Algorithm~\ref{alg:fedavg} is a FedAvg strategy found in~\cite{pmlr-v54-mcmahan17a}. 
The key parameters of this FedAvg algorithm are~$K$ 
(the number of clients in each federated learning round), 
$E$ (the number of local training epochs), 
$B$ (the local minibatch size), and~$\eta$ (the learning rate).

\begin{figure}[!htb]
\algdef{SE}[SUBALG]{Indent}{EndIndent}{}{\algorithmicend\ }%
\algtext*{Indent}
\algtext*{EndIndent}
\centering
\begin{minipage}{0.85\textwidth}
\begin{algorithm}[H]
\caption{FedAvg}\label{alg:fedavg} 
\small
\begin{algorithmic} 
\State \bbb{\texttt{/\!/} $K$ clients indexed by $k$}
\State \bbb{\texttt{/\!/} $\mathcal{P}_k$ is training dataset on client $k$}
\State \bbb{\texttt{/\!/} $n_k=|\mathcal{P}_k|$ and $n=\sum_{k=1}^K n_k$}
\State \bbb{\texttt{/\!/} $B$ is local minibatch size}
\State \bbb{\texttt{/\!/} $E$ is the number of local epochs}
\State \bbb{\texttt{/\!/} $\eta$ is the learning rate}
\State \bbb{\texttt{/\!/} $\ell(w; b)$ is local loss function evaluated on weights~$w$ 
	and minibatch~$b_{\vphantom{\sum_{M_M}}}$}
%
\State \textbf{Server Executes}:
\Indent
   \State initialize $w_0$
   \For{each round $t = 0, 1, 2, \dots$}
     \For{each client $k \in K$ \textbf{in parallel}} \bbb{\texttt{/\!/} all clients update model}
       \State $w_{t+1}^k \gets \Call{ClientUpdate}{k, w_t}$ 
     \EndFor
     \State $w_{t+1} \gets \displaystyle\sum_{k=1}^K \frac{n_k}{n} w_{t+1}^k$ \bbb{\texttt{/\!/} weighted average}
   \EndFor
\EndIndent
\Function{ClientUpdate}{$k, w$} \bbb{\texttt{/\!/} runs on client $k$}
  \State $\mathcal{B} \gets \mbox{(split $\mathcal{P}_k$ into minibatches of size $B$)}$
  \For{each local epoch $i$ from $1$ to $E$}
    \For{each minibatch $b \in \mathcal{B}$}
      \State $w \gets w - \eta \nabla \ell(w; b)$
    \EndFor
  \EndFor
  \State \textbf{return} $w$ to server
\EndFunction
\end{algorithmic}
\end{algorithm}
\end{minipage}%
\end{figure}

While there are numerous potential threats to FL systems, in this paper,
we focus on a simple label-flipping attack. That is, we specify a percentage of 
adversarial clients and a percentage of labels flipped.
Each of the adversarial clients then
flips the specified percentage of labels in its local dataset, which has the 
effect of corrupting its model update to the centralized server. 

Next, we introduce each of the seven FL models considered in this paper.
These models include examples of both classic machine learning and
neural network-based models.

\subsection{Multinominal Logistic Regression}

Multinomial Logistic Regression (MLR) is used to predict the probability of a certain category, 
where the dependent variable can represent multiple categories. It calculates the weighted sum of the independent variables 
and their respective coefficients to find the log odds---the model multiplies the value of each independent variable 
by its coefficient and adds all such values. The softmax function is then used to convert these log odds into 
probabilities for each category~\cite{MLR}.

\subsection{Support Vector Classifier}

Support Vector Machines (SVM) can be used for classification and regression. The goal when training an SVM
is to construct a hyperplane that serves as a decision boundary to separate two classes, while
maximizes the margin, which is defined as the minimum distance between the data points and the separating hyperplane.
By maximizing the margin, an SVM minimizes the chance of incorrectly classifying data points not in the training set. 
Nonlinear decision boundaries can be constructed when training an SVM by use of the so-called kernel trick, which
allows for the data to be embedded in a higher dimensional space. By carefully selecting the kernel
function, the computational complexity is minimized. Support Vector Classifiers (SVC) generalize the SVM
approach to multiclass data~\cite{Stamp_2022}.

\subsection{Multilayer Perceptron}

Multilayer Perceptrons (MLPs) are the most basic type of feedforward neural network, and they
are frequently used for supervised learning tasks. An MLP includes an input layer, one or more hidden layers, 
and an output layer. Each layer consists of multiple neurons, which are fully connected to the neurons 
in the preceding and succeeding layers~\cite{FL_MLP_intro}.

The input layer receives raw data and passes it to the first hidden layer. Hidden layers
are responsible for extracting information and learning complex patterns and features from 
the input data. Each node in these layers performs a nonlinear transformation on the weighted 
sum of its inputs, employing activation functions such as ReLU, sigmoid, or tanh. This nonlinearity 
enables the MLP to model intricate, nonlinear relationships within the data.

\subsection{Convolution Neural Network}

Convolutional Neural Networks (CNNs) are a type of feedforward neural network that specializes 
in grid-like data and, in particular, images. CNNs are optimized for dealing with local structure, as
opposed to, say, MLPs, which can effectively deal with global structure, but are too inefficient for 
complex images~\cite{Stamp_2022}.

A typical CNN includes an input layer, convolution layers, pooling layers, 
and an output layer. This neural network does not require manual feature engineering,
as it autonomously extracts features, further increasing efficiency.

%

\subsection{Random Forest}

Random Forests (RF) are ensemble learning methods widely used for classification and regression related tasks. 
An RF consists of multiple decision trees, each of which is trained on a subset of the features and data,
with a simple voting scheme typically used for classification.
Such an approach reduces overfitting and improves the generalizability of the model~\cite{RF}.

\subsection{XGBoost}

Boosting a generic learning technique that builds a strong classifier from a collection of weak classifiers.
Extreme Gradient Boosting (XGBoost) is a robust boosting technique that has performed well in many
machine learning contests~\cite{XGB}. Like Random Forest, 
our implementation of XGBoost is based on simple decision trees.

\subsection{Long Short-Term Memory} 

Long Short-Term Memory (LSTM) models represent a class of neural networking architectures 
designed to deal with sequential data. LSTMs are highly specialized types of RNNs 
that allow for long-term dependencies in the data. LSTMs mitigate the vanishing and exploding gradient
issues that plague generic RNNs, thereby enabling LSTMs to 
``remember'' information over an extended period of time, which can improve the accuracy of 
predictions~\cite{LSTM}.

\subsection{Related Work}

In this section, we briefly review previous work involving attacks on FL systems. For a more detailed 
discussion of the FL literature, see the literature review in the companion paper~\cite{Rohit}.

There exists a surprisingly large number of survey (and similar) papers dealing with attacks on FL 
systems, including~\cite{bouacida2021vulnerabilities,FL_attack_survey,jere2020taxonomy,
kumar2023impact,FL_attacks,nair2023robust,rodriguez2023survey}, among others.
These survey-like papers tend to have a broad focus, and many place an emphasis on 
categorizing the various types of
attacks that can occur at different stages of the FL process. The label-flipping 
attacks considered in this paper are considered to be
examples of poisoning attacks~\cite{tolpegin2020data}.

There is also no shortage of research papers dealing with label-flipping attacks on FL systems. 
Examples of such papers include~\cite{JDSB,JEBREEL2024111,10054157,li2021detection,LCZRWCLCW},
among many others.
However, these papers tend to be focused on the problem of detecting label-flipping attacks, 
as opposed to analyzing the effectiveness of such attacks. 
In contrast, our research is narrowly focused on the effectiveness of label-flipping attacks, as a function of the
number of adversarial clients and the percentage of labels flipped by each of the adversarial clients. 

The paper~\cite{TTGL} is an example research
into label-flipping attack effectiveness in FL. However, in~\cite{TTGL} 
the emphasis is on targeted attacks, while the research presented in this paper
does not consider targeted attack scenarios.
To the best of the authors' knowledge, there is a relative paucity of research papers that analyze 
label-flipping attack effectiveness, and we are not aware of any research that considers 
the specific problem analyzed in this paper.

\section{Implementation}\label{chap:3}

In this section, we first discuss the dataset used for our experiments. Then in the remainder of
this section, we outline our experimental design.

\subsection{Dataset}

\begin{figure}[!htb]
\centering
\includegraphics[scale=0.5]{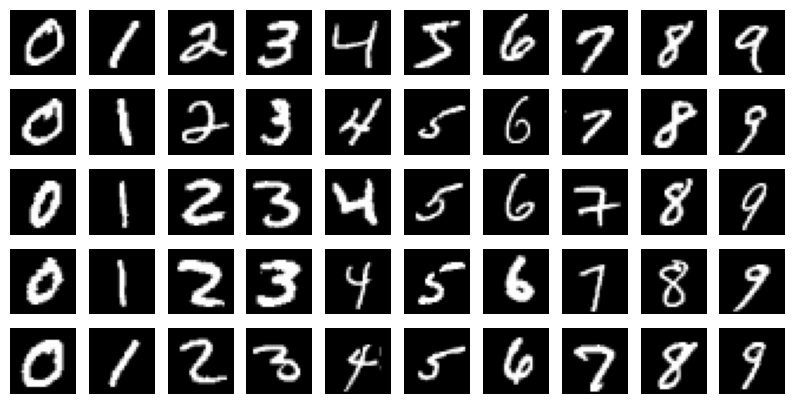}
\caption{Examples of MNIST images}
\label{fig:mnist}
\end{figure}

For all of our experiments, we use the popular MNIST dataset~\cite{lecun-mnisthandwrittendigit-2010},
which consists of handwritten digits, 0 through~9. The MNIST dataset 
is often used as a benchmark for various learning algorithms. 
This dataset consists of~60,000 training samples and~10,000 test samples. 
All samples are in the form of grayscale images of size~$28\times 28$,
with each pixel value in the range of~0 to~255, where~0 represents black and~255 represents white. 
Examples of images from the dataset are given in Figure~\ref{fig:mnist}. 

\begin{figure}[!htb]
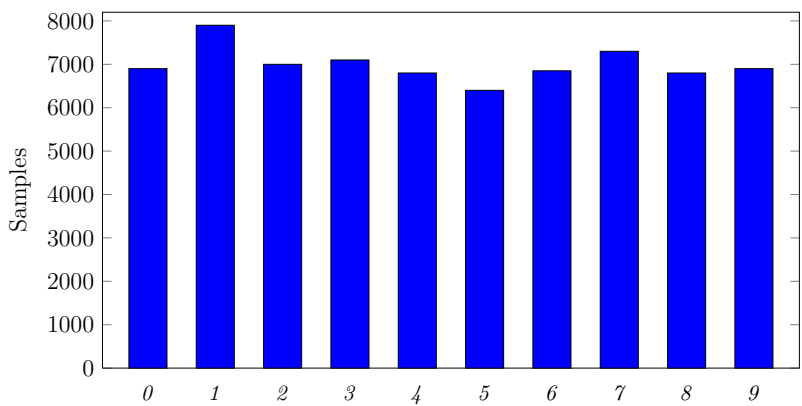

    \centering
    \input figures/barMNIST.tex
    \caption{Class distribution in MNIST dataset}\label{fig:class_distr}
\end{figure}

The MNIST dataset is approximately balanced. The precise
number of samples in each class is given in the form of a bar graph in 
Figure~\ref{fig:class_distr}.

As a preprocessing step, the MNIST images are converted into tensors or numpy arrays, 
depending on the libraries used for the specific classifier. 
The pixel values in the MNIST dataset have a mean of~1.307 and a standard deviation of~0.3081,
and these values are normalized to have a mean of~0 and a standard deviation of~1, 
which is standard practice in machine learning. 

\subsection{Experimental Design}

All federated models were trained using 
Flower: A Friendly Federated Learning Framework~\cite{beutel2020flower},
which is a Python library designed for 
such models. The \texttt{torch.utils.data.random\_split} function in Pytorch~\cite{pytorch}
was used to split the data between all the clients, and the label flipping
occurred throughout all round---in the terminology of the paper~\cite{Rohit},
we consider the FULL case.

\begin{figure}[!htb]
\centering
\includegraphics[scale=0.925]{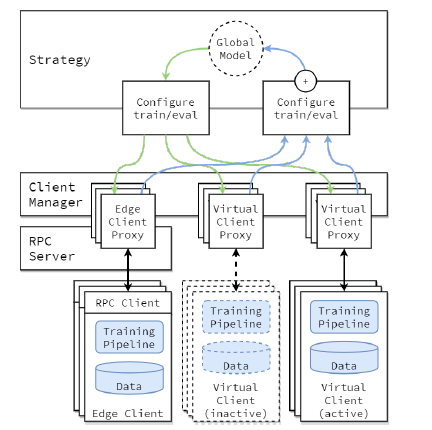}
\caption{Flower federated ML framework \cite{beutel2020flower}}
\label{fig:flower-arch}
\end{figure}

The FL stack developed for this research
has three main components, namely, the Server, Client, and Strategy.  
\begin{itemize}
\item {\bf Server}: The Server is responsible for global computations,
including aggregating the model weights, selecting the input parameters for the models, 
and sampling random clients for each FL round. 
\item {\bf Client}: The Client is responsible for executing local computations, including running 
the ML model for a set amount of epochs. The client has access to the actual data used for training and evaluation 
of model parameters.
\item {\bf Strategy}: The framework provides a Strategy abstraction which includes the logic for client selection, 
configuration, parameter aggregation, and model evaluation. Outlier detection has been implemented in this 
strategy as a defense mechanism to reject model updates from malicious clients, and is executed on the server. 
A high-level abstraction of the Flower FL framework is provided in Figure~\ref{fig:flower-arch}. 
\end{itemize}

For our experiments, the FedAvg~\cite{pmlr-v54-mcmahan17a} strategy was used to 
aggregate model weights for all models, except that a bagging aggregation 
strategy~\cite{FlowerAIFedXgbBagging2023} 
was used for aggregating model updates for tree-based models (Random Forest and XGBoost).  
Note that the clients and the server communicate through Remote Procedure Calls (RPC). 

Each experiment was performed for~10 federated rounds and the hyperparameters were adjusted 
accordingly. For example, if a model requires~120 epochs for convergence, the number of local epochs 
is set to~12 in each FL round so that at the end of the FL process, the models would have been trained for 
a total of~120 epochs.

\section{Experiments and Results}\label{chap:4}

We first consider a series of experiments where there are no adversarial clients.
These experiments serve to determine the hyperparameters for our models, and
to set baselines for accuracy. We then consider the effect of adversarial clients
on each federated model, and we conclude this section with an analysis
of the relationship between the label-flipping percentage
and the percentage of adversarial clients.

\subsection{Baseline Experiments}\label{sect:base}

Table~\ref{tab:params10} in Appendix~\ref{app:a} lists the hyperparameters tested (via grid search) for 
each model in the case of~10 clients, with
the hyperparameters selected for the best model given in boldface.
Table~\ref{tab:params100} in Appendix~\ref{app:a} contains the analogous results for
each model in the case of~100 clients.

\begin{figure}[!htb]
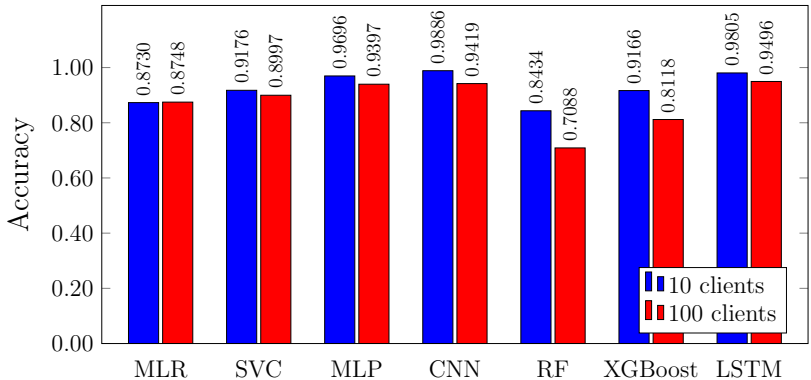

\centering
\adjustbox{scale=0.95}{
\input figures/bar_10_100.tex
}
\caption{Baseline model accuracies for~10 and~100 clients}\label{fig:acc_10_100}
\end{figure}

Figure~\ref{fig:graphs} in Appendix~\ref{app:a}
shows the accuracies for each model as a function of the number of clients,
where the number of clients ranges from~10 to~100. 
In Figure~\ref{fig:acc_10_100}, we give the accuracies for~10
and~100 clients for each model.
We observe that with the exception of MLR, all of the models perform worse as the number of
clients increases. In some cases, the degradation in accuracy for larger numbers of
clients is small (e.g., SVC and MLP), while for other models, the decline is 
more substantial (e.g., Random Forest and XGBoost).

\subsection{Adversarial Attack Experiments}\label{sect:aa}

In this section, we consider label-flipping attacks on each of the seven models. Using our best model for~10 
clients---as determined in Section~\ref{sect:base}, above---we vary the percentage of adversarial clients 
from~10\%\ to~100\%, in steps of~10\%. For each of these~10
test cases, we vary the label-flipping percentage of each adversarial client from~10\%\ to~100\%, 
again with a steps size of~10\%. This gives us~100 accuracy results for each~10-client model. We then
repeat this entire process for each model, but with~100 clients instead of~10.

The results for each model for our~10-client adversarial attack experiments are summarized 
in the form of 3-dimensional surface plots in Figure~\ref{fig:3d_10} in Appendix~\ref{app:b}. 
The corresponding results for our~100-client experiments appear in 
Figure~\ref{fig:3d_100} in Appendix~\ref{app:b}. Next,
we provide brief comments on our adversarial attack results for each of 
the seven federated models under consideration.

\subsubsection{Multinominal Logistic Regression}

From Figure~\ref{fig:graphs}(a) we observe that the MLR model accuracy drops only slightly when used
in a federated mode, as compared to an MLR model with no federated clients. However, the accuracy
of the federated MLR is fairly constant, regardless of the number of federated clients.

Comparing the attack results in Figures~\ref{fig:3d_10}(a) and~\ref{fig:3d_100}(a), we observe similar behavior.
This is not surprising, given that the MLR model performs similarly over a wide range
of federated clients.
 
\subsubsection{Support Vector Classifier}

In Figure~\ref{fig:graphs}(b) we observe that the SVC model accuracy
only decreases slightly as the number of federated clients increases.
This is similar behavior as was observed for the MLR model, above

Comparing the adversarial attacks 
in Figures~\ref{fig:3d_10}(b) and~\ref{fig:3d_100}(b), we observe that the SVC model
is highly symmetric about the $(x,y)$-plane, as compared to the other models.

\subsubsection{Multilayer Perceptron}

From Figure~\ref{fig:graphs}(c) we observe that the MLP model achieves high accuracy and,
similar to the MLR and SVC models, the accuracy does not drop significantly as
more federated clients are added.

Comparing the label-flipping attacks in Figures~\ref{fig:3d_10}(c) and~\ref{fig:3d_100}(c), 
we observe that in the 10-client case the accuracy drops precipitously, while
this drop off is somewhat smoother in the 100-client case. These graphs are the least
symmetric of the attack graphs considered so far.

\subsubsection{Convolution Neural Network}

From Figure~\ref{fig:graphs}(d) we observe that the CNN model accuracy actually improves slightly
in a federated mode, as compared to the case with no federated clients.
However, as we add more federated clients, the model accuracy degrades
much more rapidly than the three models considered above.

For the 10-client case, the attack graph in Figures~\ref{fig:3d_10}(d) for the CNN model
is similar to that of the MLP model. However, CNN 100-client case in Figure~\ref{fig:3d_100}(d)
is more erratic than any of the other models, which would seem to indicate that this model
is somewhat unstable with~100 clients.

%
%

\subsubsection{Random Forest}

From Figure~\ref{fig:graphs}(e) we observe that the Random Forest model achieves high accuracy 
when no federated clients are considered. Also,
the accuracy drop consistently as more federated clients are included,
and with~100 clients, the model performs poorly.

Comparing the label-flipping attacks in Figures~\ref{fig:3d_10}(e) and~\ref{fig:3d_100}(e), 
we observe that the qualitative behavior is similar, but the 10-client experiments
show consistently higher accuracies.
This is not too surprising, given that the decline in accuracy 
as more federated clients are used, as noted above.

\subsubsection{XGBoost}

From Figure~\ref{fig:graphs}(f) and the label-flipping attack graphs 
in Figures~\ref{fig:3d_10}(f) and~\ref{fig:3d_100}(f), 
we observe that behavior of the XGBoost models are similar to those
of the Random Forest. This is not surprising, given that they are both 
tree-based algorithms.

\subsubsection{Long Short-Term Memory} 

From Figure~\ref{fig:graphs}(g) we observe that the LSTM model achieves 
high accuracy. We also note that the accuracy of the LSTM model
drops only slightly as the number of clients increases from~10 to~100.

Comparing the label-flipping attacks in Figures~\ref{fig:3d_10}(g) and~\ref{fig:3d_100}(g), 
we observe that the LSTM behaves most similar to the MLP model. This is somewhat surprising,
since these models are dramatically different.

\subsection{Dominance Graphs}\label{sect:dom}

For a given overall level of labels flipped, the flipping can be dominated by the number of adversarial clients,
or by the percentage of labels flipped. For example, suppose that~40\%\ of the clients are adversarial,
and that each of these flips~20\%\ of the labels in their local dataset. Since the local datasets are all of the 
same size, this implies that for the model as a whole, 8\%\ of the labels are flipped. On the other hand,
if only~20\%\ of the clients are malicious, but each flips~40\%\ of the labels in their local dataset,
this also represents a case where~8\%\ of the labels are flipped. All of the 3-dimensional accuracy graphs in 
Figures~\ref{fig:3d_10} and~\ref{fig:3d_100} that are not symmetric with respect to the~$(x,y)$-plane
will---for a given level of label-flipping and selected region of the domain---perform better for one of these two cases, 
that is, the case where the percentage of adversarial clients dominates or where the percentage of 
labels flipped dominates.

To obtain better insight into this relationship between the relative percentage of labels flipped and 
the percentage of adversarial clients, we generate 2-dimensional ``dominance curves'' for each of
the seven models under consideration. Let~$c$ be the fraction of adversarial clients 
and let~$\ell$ be the fraction of labels flipped by each adversarial client. As discussed 
in Section~\ref{sect:aa}, above, we have~$c, \ell \in\{0.1,0.2,\ldots,1.0\}$,
and for each of these~100 cases we test the model and determine the accuracy. For a given model~$m$,
denote the accuracy for a specified~$c$ and~$\ell$ as~$A_m(c,\ell)$.
Note that the values~$A_m(c,\ell)$ are derived from the same experimental results 
that were used to construct the 3-dimensional surface plots in 
Figures~\ref{fig:3d_10} and~\ref{fig:3d_100} in Appendix~\ref{app:b}.

To construct the dominance curves for a given model, we consider all~100 test cases,
and whenever~$c > \ell$, then~$(c\ell,A_m(c,\ell))$ is a point on the client-dominated curve and, on the other hand,
whenever~$\ell > c$, then~$(c\ell,A_m(c,\ell))$ is a point on the flipping-dominated curve. We ignore the cases
where~$\ell$ and~$c$ are equal, since neither dominates the other.
We refer to~$c\ell$ as the label-flipping rate, since it gives the overall fraction of labels flipped.
Based on the data used to construct the 3-dimensional accuracy
graphs in Figure~\ref{fig:3d_10} or Figure~\ref{fig:3d_100}, dominances
graphs for each model appear in Figure~\ref{fig:flip} in Appendix~\ref{app:b}.\footnote{For 
each model, we selected either the~10-client or~100-client case to draw the corresponding dominance
graph in Figure~\ref{fig:flip}, depending on which of the 3-dimensional attack graphs in 
Figures~\ref{fig:3d_10} and~\ref{fig:3d_100} produced visually smoother results.} 

From Figure~\ref{fig:flip} in Appendix~\ref{app:b}, 
we observe that for the MLR model,
it is more effective---from an attacker's perspective---to have fewer clients 
flipping a higher percentage of labels. In contrast,
for the MLP model---and to a lesser extent, the CNN model---for a given 
overall percentage of labels flipped,
a stronger attack will consist of more adversarial clients,
each flipping a smaller fraction of the labels.
In fact, each model has a bias (or biases, depending on the label-flipping rate)
towards client dominance or flipping dominance, with the exception of SVC, which is
essentially unbiased in this respect throughout the entire range of label-flipping. 

The insights provided by the graphs in Figure~\ref{fig:flip} in Appendix~\ref{app:b}
could be used to help determine a preferred model, based on the likelihood of various attack scenarios. 
For example, if there is a higher probability
that many adversarial clients will send slightly corrupted updates, we would prefer different models
as compared to the case where our primary concern is relatively few 
adversarial clients, with each potentially sending relatively highly-corrupted updates.
As another example, we might have stronger defenses against specific types of attacks,
in which case we could choose models that are inherently more robust against the types of attacks
that are more difficult to detect.

\section{Conclusion}\label{chap:5}

In this paper, we empirically  analyzed label-flipping attacks against the following
federated learning models:
Multinominal Logistic Regression (MLR),
Support Vector Classifier (SVC),
Multilayer Perceptron (MLP),
Convolution Neural Network (CNN),
Random Forest,
XGBoost,
and
Long Short-Term Memory (LSTM).
We found that all model have reduced accuracy as more clients are added, although for most models,
the reduction was small within the range of~10 to~100 clients. We then considered the 10-client and
100-client cases in more detail, graphing the accuracy as a function of the percentage of 
adversarial clients and the percentage of labels flipped by each adversarial client. We then further analyzed 
the relationship between the percentage of adversarial clients and the percentage
of labels flipped. For a given overall percentage of labels flipped, we found that some
models are inherently more robust when there are fewer adversarial clients
flipping a higher percentage of labels, whereas other models were more robust
in the case where there are more adversarial clients, but each flips a smaller
percentage of the labels in their local dataset. This has practical implications, as
we might, for example, choose models that are more robust against likely adversarial
attacks, or we might choose models that are more robust against attacks that
are harder to defend against.

For future work, it would be interesting to extend the research in this paper to other FL models.
We could also consider more fine-grained attack scenarios, with smaller steps
in the percentage of adversarial clients and the percentage of labels flipped.
It would be worthwhile to consider more sophisticated adversarial attacks
involving strategies other simple label-flipping. Targeted attacks would be
interesting, where the goal is to maintain the overall accuracy, but to force the 
misclassification of samples belonging to a specific class. Of course, an empirical analysis 
of the effectiveness of various defensive strategies would be another interesting line of research.

\bibliographystyle{plain}
\bibliography{references.bib, references2.bib}



\titleformat{\section}{\normalfont\large\bfseries}{}{0em}{#1\ \thesection}
\setcounter{section}{0}
\renewcommand{\thesection}{\Alph{section}}
\renewcommand{\thesubsection}{A.\arabic{subsection}}
\setcounter{table}{0}
\renewcommand{\thetable}{A.\arabic{table}}
\setcounter{figure}{0}
\renewcommand{\thefigure}{A.\arabic{figure}}
\section{Appendix}\label{app:a}

Table~\ref{tab:params10} lists the hyperparameter values tested for each of the federated learning
models, in the case where there are~10 clients. The values selected are given in boldface.
Table~\ref{tab:params100} lists the corresponding hyperparameter values tested and
selected for the federated learning models in the case where there are~100 clients.
In Figure~\ref{fig:graphs}, we graph the accuracy of each federated model
as a function of the number of clients.

\begin{table}[!htb]
\centering
\caption{Hyperparameters (10 clients)}\label{tab:params10}
\adjustbox{scale=0.75}{
\begin{tabular}{c|cc|cc}
\toprule
\multirow{2}{*}{Model} 
 & \multirow{2}{*}{Hyperparameters} & \multirow{2}{*}{Tested values} & \multicolumn{2}{c}{Accuracy} \\
& & & Train & Test \\
\midrule
\multirow{6}{*}{MLR} 
& \texttt{learning\_rate} & $\left[\textbf{0.01}, 0.0001\right]$ & \multirow{6}{*}{0.8715} & \multirow{6}{*}{0.8730} \\
& \texttt{batch\_size} & $\left [\textbf{20}, 64, 128\right]$ & & \\
& Epochs & $\left [\textbf{1}, 10, 20\right]$ & & \\
& momentum & \textbf{0.9} & & \\
& \texttt{penalty} & \textbf{l2} & & \\
& \texttt{warm\_start} & \textbf{True} & & \\
\midrule 
\multirow{5}{*}{SVC} 
& \texttt{learning\_rate} & $\left[\textbf{0.01}, 0.001\right]$ & \multirow{5}{*}{0.9026} & \multirow{5}{*}{0.9176} \\
& \texttt{batch\_size} & $\left[\textbf{20}, 64, 128\right]$ & & \\
& Epochs & $\left[\textbf{1}, 10, 20\right]$ & & \\
 & \texttt{momentum} & \textbf{0.9} & &\\
 & \texttt{penalty} & \textbf{l2} & &\\
\midrule 
\multirow{5}{*}{MLP} 
& \texttt{learning\_rate} & $\left[\textbf{0.003}, 0.0001\right]$ & \multirow{5}{*}{0.9793} & \multirow{5}{*}{0.9696} \\
& \texttt{batch\_size} & $\left [\textbf{20}, 64, 128\right]$ & & \\
& Epochs & $\left [\textbf{1}, 10, 20\right]$ & & \\
& Optimizer & $\left[\textbf{Adam}, \mbox{RMSProp}\right]$ & & \\
& \texttt{image\_dim} & $\left[\textbf{128}, 256\right]$ & & \\
\midrule 
\multirow{5}{*}{CNN} 
& \texttt{learning\_rate} & $\left[\textbf{0.01}, 0.001\right]$ & \multirow{5}{*}{0.9896} & \multirow{5}{*}{0.9886} \\
& \texttt{batch\_size} & $\left [\textbf{20}, 64, 128 \right]$ & & \\
& Epochs & $\left [\textbf{1},10,20\right]$ & & \\
& Optimizer & $\left[\textbf{Adam}, \mbox{RMSProp}\right]$ & & \\
& \texttt{image\_dim} & $\left[\textbf{128}, 256, 375\right]$ & & \\
\midrule 
\multirow{11}{*}{Random Forest} 
& \texttt{learning\_rate} & $\left[\textbf{0.08}, 0.0001\right]$ & \multirow{10}{*}{0.8471} & \multirow{10}{*}{0.8434} \\
& \texttt{num\_parallel\_tree} & $\left [32, \textbf{100}, 128\right]$ & & \\
& \texttt {max\_depth} & $\left[2, 4, \textbf{6}\right]$ & & \\
& \texttt{Epochs} & $\left [\textbf{1}, 10, 20\right]$& & \\
& \texttt{colsample\_bytree} & $\left[\textbf{0.963}, 0.7 \mbox{0.5}\right]$ & & \\
& \texttt{subsample} & $\left[\textbf{0.97}, 0.7, 0.5\right]$ & & \\
& \texttt {objective} & \textbf{multi:softmax} & & \\
& \texttt {eval\_metric} & \textbf{mlogloss} & & \\
& \texttt {alpha} & $\left[2, 4, \textbf{8}\right]$ & & \\
& \texttt {Lambda} & $[1, \textbf{2}, 3]$ & & \\
& \texttt {tree\_method} & \textbf{hist} & & \\
\midrule 
\multirow{10}{*}{XGBoost} 
& \texttt{learning\_rate} & $\left[0.001, \textbf{0.08}\right]$ & \multirow{10}{*}{0.9044} & \multirow{10}{*}{0.9166} \\
& \texttt{local\_epochs} & $\left [\textbf{1}, 10, 20\right]$& & \\
& \texttt{max\_depth} & $\left [\textbf{6}, 10, 12\right]$& & \\
& \texttt{subsample} & $\left [0.50, 0.75, \textbf{1}\right]$& & \\
& \texttt{colsample\_bytree} & $\left [0.50, 0.75, \textbf{1}\right]$& & \\
& \texttt{objective} & \textbf{multi:softmax} & & \\
& \texttt{eval\_metric} & \textbf{mlogloss} & & \\
& \texttt{alpha} & $\left [2, 4, \textbf{8}\right]$& & \\
& \texttt{Lambda} & $\left [2, \textbf{4}, 8\right]$& & \\
& \texttt{tree\_method} & \textbf{hist} & & \\
\midrule
\multirow{4}{*}{LSTM} 
& \texttt{learning\_rate} & $\left[\textbf{0.001}, 0.1\right]$ & \multirow{4}{*}{0.9906} & \multirow{4}{*}{0.9805} \\
& \texttt{batch\_size} & $\left [24, \textbf{64}, 128 \right]$ & & \\
& Epochs & $\left [1, \textbf{10}, 20\right]$ & & \\
& Optimizer & $\left[\textbf{SGD}, \mbox{Adam}\right]$ & & \\
\bottomrule 
\end{tabular}
}
\end{table}

\begin{table}[!htb]
\centering
\caption{Hyperparameters (100 clients)}\label{tab:params100}
\adjustbox{scale=0.75}{
\begin{tabular}{c|cc|cc}
\toprule
\multirow{2}{*}{Model} 
 & \multirow{2}{*}{Hyperparameters} & \multirow{2}{*}{Tested values} & \multicolumn{2}{c}{Accuracy} \\
& & & Train & Test \\
\midrule
\multirow{6}{*}{MLR} 
& \texttt{learning\_rate} & $\left[\textbf{0.001}, 0.0001\right]$ & \multirow{6}{*}{0.8537} & \multirow{6}{*}{0.8748} \\
& \texttt{batch\_size} & $\left [\textbf{20}, 64, 128\right]$ & & \\
& Epochs & $\left [\textbf{1}, 10, 20\right]$ & & \\
& momentum & \textbf{0.9} & & \\
& \texttt{penalty} & \textbf{l2} & & \\
& \texttt{warm\_start} & \textbf{True} & & \\
\midrule 
\multirow{5}{*}{SVC} 
& \texttt{learning\_rate} & $\left[\textbf{0.01}, 0.001\right]$ & \multirow{5}{*}{0.8983} & \multirow{5}{*}{0.8997} \\
& \texttt{batch\_size} & $\left[\textbf{20}, 64, 128\right]$ & & \\
& Epochs & $\left[\textbf{1}, 10, 20\right]$ & & \\
 & \texttt{momentum} & \textbf{0.9} & &\\
 & \texttt{penalty} & \textbf{l2} & &\\
\midrule 
\multirow{5}{*}{MLP} 
& \texttt{learning\_rate} & $\left[\textbf{0.003}, 0.0001\right]$ & \multirow{5}{*}{0.9524} & \multirow{5}{*}{0.9397} \\
& \texttt{batch\_size} & $\left [\textbf{20}, 64, 128\right]$ & & \\
& Epochs & $\left [\textbf{1}, 10, 20\right]$ & & \\
& Optimizer & $\left[\textbf{Adam}, \mbox{RMSProp}\right]$ & & \\
& \texttt{image\_dim} & $\left[\textbf{128}, 256\right]$ & & \\
\midrule 
\multirow{5}{*}{CNN} 
& \texttt{learning\_rate} & $\left[\textbf{0.01}, 0.001\right]$ & \multirow{5}{*}{0.9317} & \multirow{5}{*}{0.9419} \\
& \texttt{batch\_size} & $\left [\textbf{20}, 64, 128 \right]$ & & \\
& Epochs & $\left [\textbf{1},10,20\right]$ & & \\
& Optimizer & $\left[\textbf{Adam}, \mbox{RMSProp}\right]$ & & \\
& \texttt{image\_dim} & $\left[\textbf{128}, 256, 375\right]$ & & \\
\midrule 
\multirow{11}{*}{Random Forest} 
& \texttt{learning\_rate} & $\left[\textbf{0.08}, 0.0001\right]$ & \multirow{10}{*}{0.7157} & \multirow{10}{*}{0.7088} \\
& \texttt{num\_parallel\_tree} & $\left [32, \textbf{100}, 128\right]$ & & \\
& \texttt {max\_depth} & $\left[2, 4, \textbf{6}\right]$ & & \\
& \texttt{Epochs} & $\left [\textbf{1}, 10, 20\right]$& & \\
& \texttt{colsample\_bytree} & $\left[\textbf{0.963}, 0.7 \mbox{0.5}\right]$ & & \\
& \texttt{subsample} & $\left[\textbf{0.97}, 0.7, 0.5\right]$ & & \\
& \texttt {objective} & \textbf{multi:softmax} & & \\
& \texttt {eval\_metric} & \textbf{mlogloss} & & \\
& \texttt {alpha} & $\left[2, 4, \textbf{8}\right]$ & & \\
& \texttt {Lambda} & $[1, \textbf{2}, 3]$ & & \\
& \texttt {tree\_method} & \textbf{hist} & & \\
\midrule 
\multirow{10}{*}{XGBoost} 
& \texttt{learning\_rate} & $\left[0.001, \textbf{0.08}\right]$ & \multirow{10}{*}{0.8001} & \multirow{10}{*}{0.8118} \\
& \texttt{local\_epochs} & $\left [\textbf{1}, 10, 20\right]$& & \\
& \texttt{max\_depth} & $\left [\textbf{6}, 10, 12\right]$& & \\
& \texttt{subsample} & $\left [0.50, 0.75, \textbf{1}\right]$& & \\
& \texttt{colsample\_bytree} & $\left [0.50, 0.75, \textbf{1}\right]$& & \\
& \texttt{objective} & \textbf{multi:softmax} & & \\
& \texttt{eval\_metric} & \textbf{mlogloss} & & \\
& \texttt{alpha} & $\left [2, 4, \textbf{8}\right]$& & \\
& \texttt{Lambda} & $\left [2, \textbf{4}, 8\right]$& & \\
& \texttt{tree\_method} & \textbf{hist} & & \\
\midrule
\multirow{4}{*}{LSTM} 
& \texttt{learning\_rate} & $\left[\textbf{0.001}, 0.1\right]$ & \multirow{4}{*}{0.9503} & \multirow{4}{*}{0.9496} \\
& \texttt{batch\_size} & $\left [24, \textbf{64}, 128 \right]$ & & \\
& Epochs & $\left [1, \textbf{10}, 20\right]$ & & \\
& Optimizer & $\left[\textbf{SGD}, \mbox{Adam}\right]$ & & \\
\bottomrule 
\end{tabular}
}
\end{table}

\begin{figure}[!htb]
\centering
\begin{tabular}{ccc}
\input figures/graph_MLR.tex
& \ \ \ \ &
\input figures/graph_SVC.tex
\\[-0.5ex]
\adjustbox{scale=0.85}{(a) MLR} & & \adjustbox{scale=0.85}{(b) SVC}\\ \\[-1.0ex]
\input figures/graph_MLP.tex
& &
\input figures/graph_CNN.tex
\\[-0.5ex]
\adjustbox{scale=0.85}{(c) MLP} & & \adjustbox{scale=0.85}{(d) CNN}\\ \\[-1.0ex]
\input figures/graph_RF.tex
& &
\input figures/graph_XGB.tex
\\[-0.5ex]
\adjustbox{scale=0.85}{(e) Random Forest} & & \adjustbox{scale=0.85}{(f) XGBoost}\\ \\[-1.0ex]
\multicolumn{3}{c}{\input figures/graph_LSTM.tex}
\\[-0.5ex]
\multicolumn{3}{c}{\adjustbox{scale=0.85}{(g) LSTM}}
\end{tabular}
\caption{Accuracy as a function of the number of clients}\label{fig:graphs}
\end{figure}

\clearpage

\renewcommand{\thesubsection}{B.\arabic{subsection}}
\setcounter{table}{0}
\renewcommand{\thetable}{B.\arabic{table}}
\setcounter{figure}{0}
\renewcommand{\thefigure}{B.\arabic{figure}}
\section{Appendix}\label{app:b}

Figure~\ref{fig:flip} contains
``dominance graphs'' that emphasizes the relationship
between the percentage of labels flipped and the percentage of adversarial
clients, as discussed in Section~\ref{sect:dom}.
Figure~\ref{fig:3d_10} contains 3-dimensional graphs of label-flipping attacks for each 
federated model, in the case of~10 clients. Figure~\ref{fig:3d_100}
gives the analogous graphs for the case where there are~100 clients. 

\begin{figure}[!htb]
\centering
\begin{tabular}{ccc}
\input figures/dom_MLR.tex
& \ \ \ \ &
\input figures/dom_SVC.tex
\\[-0.5ex]
\adjustbox{scale=0.85}{(a) MLR (10 clients)} & & \adjustbox{scale=0.85}{(b) SVC (10 clients)}\\ \\[-1ex]
\input figures/dom_MLP.tex
& &
\input figures/dom_CNN.tex
\\[-0.5ex]
\adjustbox{scale=0.85}{(c) MLP (100 clients)} & & \adjustbox{scale=0.85}{(d) CNN (100 clients)}\\ \\[-1ex]
\input figures/dom_RF.tex
& &
\input figures/dom_XGB.tex
\\[-0.5ex]
\adjustbox{scale=0.85}{(e) Random Forest (10 clients)} & & \adjustbox{scale=0.85}{(f) XGBoost (10 clients)}\\ \\[-1ex]
\multicolumn{3}{c}{\input figures/dom_LSTM.tex}
\\[-0.5ex]
\multicolumn{3}{c}{\adjustbox{scale=0.85}{(g) LSTM (100 clients)}}
\end{tabular}
\caption{Accuracy as a function of the label-flipping rate}\label{fig:flip}
\end{figure}

\begin{figure}[!htb]
\centering
\advance\tabcolsep by -6pt
\begin{tabular}{cc}
\adjustbox{scale=0.55}{
\input figures/3d_10_MLR.tex
}
&
\adjustbox{scale=0.55}{
\input figures/3d_10_SVC.tex
}
\\[-0.25ex]
\adjustbox{scale=0.85}{(a) MLR} & \adjustbox{scale=0.85}{(b) SVC}\\ \\
\adjustbox{scale=0.55}{
\input figures/3d_10_MLP.tex
}
&
\adjustbox{scale=0.55}{
\input figures/3d_10_CNN.tex
}
\\[-0.25ex]
\adjustbox{scale=0.85}{(c) MLP} & \adjustbox{scale=0.85}{(d) CNN}\\ \\
\adjustbox{scale=0.55}{
\input figures/3d_10_RF.tex
}
&
\adjustbox{scale=0.55}{
\input figures/3d_10_XGB.tex
}
\\[-0.25ex]
\adjustbox{scale=0.85}{(e) Random Forest} & \adjustbox{scale=0.85}{(f) XGBoost}\\ \\
\multicolumn{2}{c}{\adjustbox{scale=0.55}{
\input figures/3d_10_LSTM.tex
}}
\\[-0.25ex]
\multicolumn{2}{c}{\adjustbox{scale=0.85}{(g) LSTM}}
\end{tabular}
\caption{Accuracy as a function of adversarial clients and label-flipping (10 clients)}\label{fig:3d_10} 
\end{figure}

\begin{figure}[!htb]
\centering
\advance\tabcolsep by -6pt
\begin{tabular}{cc}
\adjustbox{scale=0.55}{
\input figures/3d_100_MLR.tex
}
&
\adjustbox{scale=0.55}{
\input figures/3d_100_SVC.tex
}
\\[-0.25ex]
\adjustbox{scale=0.85}{(a) MLR} & \adjustbox{scale=0.85}{(b) SVC}\\ \\
\adjustbox{scale=0.55}{
\input figures/3d_100_MLP.tex
}
&
\adjustbox{scale=0.55}{
\input figures/3d_100_CNN.tex
}
\\[-0.25ex]
\adjustbox{scale=0.85}{(c) MLP} & \adjustbox{scale=0.85}{(d) CNN}\\ \\
\adjustbox{scale=0.55}{
\input figures/3d_100_RF.tex
}
&
\adjustbox{scale=0.55}{
\input figures/3d_100_XGB.tex
}
\\[-0.25ex]
\adjustbox{scale=0.85}{(e) Random Forest} & \adjustbox{scale=0.85}{(f) XGBoost}\\ \\
\multicolumn{2}{c}{\adjustbox{scale=0.55}{
\input figures/3d_100_LSTM.tex
}}
\\[-0.25ex]
\multicolumn{2}{c}{\adjustbox{scale=0.85}{(g) LSTM}}
\end{tabular}
\caption{Accuracy as a function of adversarial clients and label-flipping (100 clients)}\label{fig:3d_100} 
\end{figure}

\end{document}

%% file: preamble.tex
\usepackage{amsmath,amsthm, amsfonts, amssymb, amsxtra, amsopn}
\usepackage{pgfplots}
\pgfplotsset{compat=1.3}
\usepackage{pgfplotstable}
\usepackage{graphicx,grffile}
\usepackage{multirow}
\usepackage{algorithm} 
\usepackage[noend]{algpseudocode} 
\usepackage{longtable}
\usepackage{booktabs}
\usepackage{listings}
\usepackage{cmap}
\usepackage{colortbl}
\usepackage{adjustbox}
\usepackage{epsfig}

\usepackage[tableposition=top,font=small,skip=5pt]{caption}
\usepackage{subcaption}
\usepackage{makecell} 

\usepackage[explicit]{titlesec}

\def\bbb#1{{\color{blue}#1}}

\PassOptionsToPackage{hyphens}{url}

\usepackage{hyperref}
\hypersetup{colorlinks=true,linkcolor=black,citecolor=black,urlcolor=blue,filecolor=black}
\hypersetup{pdfpagemode=UseNone,pdfstartview=}

\usepackage{enumitem}
\setlist[itemize]{noitemsep, topsep=0pt}

\advance\oddsidemargin by -0.35in
\advance\textwidth by 0.7in

\advance\topmargin by -0.4in
\advance\textheight by 0.8in

\long\def\symbolfootnotetext[#1]#2{\begingroup%
\def\thefootnote{\fnsymbol{footnote}}\footnotetext[#1]{#2}\endgroup}

%% file: figures/barMNIST.tex
\begin{tikzpicture}[scale=0.8, every node/.style={scale=0.95}]
\pgfkeys{/pgf/number format/.cd,1000 sep={}}
\begin{axis}[
        width  = 0.85*\textwidth,
        height = 7.5cm,
        ymin=0,ymax=8200,
        ytick={0,1000,2000,3000,4000,5000,6000,7000,8000},
        major x tick style = transparent,
        ybar=5*\pgflinewidth,
        bar width=18.0pt,
        ylabel = {Samples},
        symbolic x coords={0,1,2,3,4,5,6,7,8,9},
        xticklabels={\textit{0},\textit{1},\textit{2},\textit{3},\textit{4},\textit{5},\textit{6},\textit{7},\textit{8},\textit{9}},
	y tick label style={
    		/pgf/number format/.cd,
   		fixed,
   		fixed zerofill,
    		precision=0},
        xtick = data,
        x tick label style={
		font=\small,
		},
        enlarge x limits=0.075,
        legend cell align=left,
        legend pos=south east,
]
\addplot [fill=blue,opacity=1.00]
coordinates {
(0, 6900)
(1, 7900)
(2, 7000)
(3, 7100)
(4, 6800)
(5, 6400)
(6, 6850)
(7, 7300)
(8, 6800)
(9, 6900)
};
\end{axis}
\end{tikzpicture}

%% file: figures/bar_10_100.tex
\begin{tikzpicture}[scale=0.8, every node/.style={scale=1.0}]
\pgfkeys{/pgf/number format/.cd,1000 sep={}}
\begin{axis}[
        width  = 0.9*\textwidth,
        height = 7.5cm,
        ymin=0.0,ymax=1.225,
        ytick={0.0, 0.2, 0.4, 0.6, 0.8, 1.0},
        major x tick style = transparent,
        ybar=5*\pgflinewidth,
        bar width=15.0pt,
        ylabel = {Accuracy},
        ylabel style = {scale = 1.2},
        symbolic x coords={MLR, SVC, MLP, CNN, RF, XGBoost, LSTM},
        xticklabels={MLR, SVC, MLP, CNN, RF, XGBoost, LSTM},
	y tick label style={
    		/pgf/number format/.cd,
   		fixed,
   		fixed zerofill,
    		precision=2},
        xtick = data,
        x tick label style={
		},
        nodes near coords,
        every node near coord/.append style={rotate=90, scale=0.825,
        								   anchor=west, 
								   /pgf/number format/.cd,
								   fixed,
								   fixed zerofill,
								   precision=4},
        enlarge x limits=0.10,
        legend cell align=left,
        legend pos=south east,
]
\addplot [fill=blue,opacity=1.00]
coordinates {
(MLR, 0.8730)
(SVC, 0.9176)
(MLP, 0.9696)
(CNN, 0.9886)
(RF, 0.8434)
(XGBoost, 0.91659)
(LSTM, 0.9805)
};
\addlegendentry{10 clients}
\addplot [fill=red,opacity=1.00]
coordinates {
(MLR, 0.8748)
(SVC, 0.8997)
(MLP, 0.9397)
(CNN, 0.9419)
(RF, 0.7088)
(XGBoost, 0.81176)
(LSTM, 0.9496)
};
\addlegendentry{100 clients}
\end{axis}
\end{tikzpicture}

%% file: figures/graph_MLR.tex
\begin{tikzpicture}[scale=0.475]
\begin{axis}[ 
		   width=0.7\textwidth,
		   height=0.575\textwidth,
	 	   x tick label style={scale=1.25,
			scale=0.95,
   		 	/pgf/number format/.cd,
			/pgf/number format/1000 sep={},
   			fixed,
   			fixed zerofill,
    			precision=0
		   },
		   x label style={scale=1.5},
	 	   y tick label style={scale=1.25,
    		 	/pgf/number format/.cd,
   			fixed,
   			fixed zerofill,
    			precision=2
		    },
		   y label style={scale=1.5},
                    xmin=10,xmax=100,
                    ymin=0.2,ymax=1.02,
                    xtick={10,20,30,40,50,60,70,80,90,100},
                    ytick={0.3,0.4,0.5,0.6,0.7,0.8,0.9,1.0},
                    xlabel={Number of clients},
                    ylabel={Accuracy}] 
\addplot[color=red,ultra thick,mark=none] coordinates { 
(10, 0.873)
(20, 0.8751)
(30, 0.8771)
(40, 0.8785)
(50, 0.881)
(60, 0.8733)
(70, 0.8690)
(80, 0.8627)
(90, 0.8698)
(100, 0.8761)
};
\end{axis}
\end{tikzpicture}

%% file: figures/graph_SVC.tex
\begin{tikzpicture}[scale=0.475]
\begin{axis}[ 
		   width=0.7\textwidth,
		   height=0.575\textwidth,
	 	   x tick label style={scale=1.25,
			scale=0.95,
   		 	/pgf/number format/.cd,
			/pgf/number format/1000 sep={},
   			fixed,
   			fixed zerofill,
    			precision=0
		   },
		   x label style={scale=1.5},
	 	   y tick label style={scale=1.25,
    		 	/pgf/number format/.cd,
   			fixed,
   			fixed zerofill,
    			precision=2
		    },
		   y label style={scale=1.5},
                    xmin=10,xmax=100,
                    ymin=0.2,ymax=1.02,
                    xtick={0,10,20,30,40,50,60,70,80,90,100},
                    ytick={0.3,0.4,0.5,0.6,0.7,0.8,0.9,1.0},
                    xlabel={Number of clients},
                    ylabel={Accuracy}] 

\addplot[color=red,ultra thick,mark=none] coordinates { 
(10, 0.9176)
(20, 0.9165)
(30, 0.9155)
(40, 0.9116)
(50, 0.9091)
(60, 0.9057)
(70, 0.9038)
(80, 0.9021)
(90, 0.9009)
(100, 0.8997)
};
\end{axis}
\end{tikzpicture}

%% file: figures/graph_MLP.tex
\begin{tikzpicture}[scale=0.475]
\begin{axis}[ 
		   width=0.7\textwidth,
		   height=0.575\textwidth,
	 	   x tick label style={scale=1.25,
			scale=0.95,
   		 	/pgf/number format/.cd,
			/pgf/number format/1000 sep={},
   			fixed,
   			fixed zerofill,
    			precision=0
		   },
		   x label style={scale=1.5},
	 	   y tick label style={scale=1.25,
    		 	/pgf/number format/.cd,
   			fixed,
   			fixed zerofill,
    			precision=2
		    },
		   y label style={scale=1.5},
                    xmin=10,xmax=100,
                    ymin=0.2,ymax=1.02,
                    xtick={0,10,20,30,40,50,60,70,80,90,100},
                    ytick={0.3,0.4,0.5,0.6,0.7,0.8,0.9,1.0},
                    xlabel={Number of clients},
                    ylabel={Accuracy}] 
\addplot[color=red,ultra thick,mark=none] coordinates { 
(10, 0.9696)
(20, 0.9688)
(30, 0.9595)
(40, 0.958)
(50, 0.9521)
(60, 0.9507)
(70, 0.9480885311871228)
(80, 0.9432)
(90, 0.9443443443443443)
(100, 0.9397)
};
\end{axis}
\end{tikzpicture}

%% file: figures/graph_CNN.tex
\begin{tikzpicture}[scale=0.475]
\begin{axis}[ 
		   width=0.7\textwidth,
		   height=0.575\textwidth,
	 	   x tick label style={scale=1.25,
			scale=0.95,
   		 	/pgf/number format/.cd,
			/pgf/number format/1000 sep={},
   			fixed,
   			fixed zerofill,
    			precision=0
		   },
		   x label style={scale=1.5},
	 	   y tick label style={scale=1.25,
    		 	/pgf/number format/.cd,
   			fixed,
   			fixed zerofill,
    			precision=2
		    },
		   y label style={scale=1.5},
                    xmin=10,xmax=100,
                    ymin=0.2,ymax=1.02,
                    xtick={10,20,30,40,50,60,70,80,90,100},
                    ytick={0.3,0.4,0.5,0.6,0.7,0.8,0.9,1.0},
                    xlabel={Number of clients},
                    ylabel={Accuracy}] 
\addplot[color=red,ultra thick,mark=none] coordinates { 
(10, 0.9886)
(20, 0.9826)
(30, 0.9784)
(40, 0.9763)
(50, 0.9683)
(60, 0.9657)
(70, 0.9606)
(80, 0.9553)
(90, 0.9451)
(100, 0.9419)
};
\end{axis}
\end{tikzpicture}

%% file: figures/graph_RF.tex
\begin{tikzpicture}[scale=0.475]
\begin{axis}[ 
		   width=0.7\textwidth,
		   height=0.575\textwidth,
	 	   x tick label style={scale=1.25,
			scale=0.95,
   		 	/pgf/number format/.cd,
			/pgf/number format/1000 sep={},
   			fixed,
   			fixed zerofill,
    			precision=0
		   },
		   x label style={scale=1.5},
	 	   y tick label style={scale=1.25,
    		 	/pgf/number format/.cd,
   			fixed,
   			fixed zerofill,
    			precision=2
		    },
		   y label style={scale=1.5},
                    xmin=10,xmax=100,
                    ymin=0.2,ymax=1.02,
                    xtick={10,20,30,40,50,60,70,80,90,100},
                    ytick={0.3,0.4,0.5,0.6,0.7,0.8,0.9,1.0},
                    xlabel={Number of clients},
                    ylabel={Accuracy}] 
\addplot[color=red,ultra thick,mark=none] coordinates { 
(10, 0.8434)
(20, 0.80954)
(30, 0.80969)
(40, 0.78997)
(50, 0.77022)
(60, 0.76384)
(70, 0.7552)
(80, 0.7371)
(90, 0.7123)
(100, 0.7088)
};
\end{axis}
\end{tikzpicture}

%% file: figures/graph_XGB.tex
\begin{tikzpicture}[scale=0.475]
\begin{axis}[ 
		   width=0.7\textwidth,
		   height=0.575\textwidth,
	 	   x tick label style={scale=1.25,
			scale=0.95,
   		 	/pgf/number format/.cd,
			/pgf/number format/1000 sep={},
   			fixed,
   			fixed zerofill,
    			precision=0
		   },
		   x label style={scale=1.5},
	 	   y tick label style={scale=1.25,
    		 	/pgf/number format/.cd,
   			fixed,
   			fixed zerofill,
    			precision=2
		    },
		   y label style={scale=1.5},
                    xmin=10,xmax=100,
                    ymin=0.2,ymax=1.02,
                    xtick={0,10,20,30,40,50,60,70,80,90,100},
                    ytick={0.3,0.4,0.5,0.6,0.7,0.8,0.9,1.0},
                    xlabel={Number of clients},
                    ylabel={Accuracy}] 
\addplot[color=red,ultra thick,mark=none] coordinates { 
(10, 0.91659)
(20, 0.89904)
(30, 0.87296)
(40, 0.86624)
(50, 0.8505)
(60, 0.8313)
(70, 0.8246)
(80, 0.8242)
(90, 0.81366)
(100, 0.81176)
};
\end{axis}
\end{tikzpicture}

%% file: figures/graph_LSTM.tex
\begin{tikzpicture}[scale=0.475]
\begin{axis}[ 
		   width=0.7\textwidth,
		   height=0.575\textwidth,
	 	   x tick label style={scale=1.25,
			scale=0.95,
   		 	/pgf/number format/.cd,
			/pgf/number format/1000 sep={},
   			fixed,
   			fixed zerofill,
    			precision=0
		   },
		   x label style={scale=1.5},
	 	   y tick label style={scale=1.25,
    		 	/pgf/number format/.cd,
   			fixed,
   			fixed zerofill,
    			precision=2
		    },
		   y label style={scale=1.5},
                    xmin=10,xmax=100,
                    ymin=0.2,ymax=1.02,
                    xtick={10,20,30,40,50,60,70,80,90,100},
                    ytick={0.3,0.4,0.5,0.6,0.7,0.8,0.9,1.0},
                    xlabel={Number of clients},
                    ylabel={Accuracy}] 
\addplot[color=red,ultra thick,mark=none] coordinates { 
(10, 0.9805)
(20, 0.9743)
(30, 0.9703)
(40, 0.9675)
(50, 0.9636)
(60, 0.9611)
(70, 0.9582)
(80, 0.9541)
(90, 0.9508)
(100, 0.9496)
};
\end{axis}
\end{tikzpicture}

%% file: figures/dom_MLR.tex
\begin{tikzpicture}[scale=0.475]
\begin{axis}[ 
		   width=0.7\textwidth,
		   height=0.5\textwidth,
	 	   x tick label style={scale=1.25,
			scale=0.95,
   		 	/pgf/number format/.cd,
			/pgf/number format/1000 sep={},
   			fixed,
   			fixed zerofill,
    			precision=1
		   },
		   x label style={scale=1.5},
	 	   y tick label style={scale=1.25,
    		 	/pgf/number format/.cd,
   			fixed,
   			fixed zerofill,
    			precision=1
		    },
		   y label style={scale=1.5},
                    xmin=0.0,xmax=0.92,
                    ymin=0.0,ymax=1.02,
                    xtick={0.0,0.1,0.2,0.3,0.4,0.5,0.6,0.7,0.8,0.9},
                    ytick={0.1,0.2,0.3,0.4,0.5,0.6,0.7,0.8,0.9,1.0},
                    xlabel={Label flipping rate},
                    ylabel={Accuracy},
                    legend cell align=left,
                    legend pos=north east,
                    ] 
\addplot[color=red,ultra thick,mark=none] coordinates { 
(0.020000, 0.873600)
(0.030000, 0.875500)
(0.040000, 0.875100)
(0.050000, 0.879600)
(0.060000, 0.870950)
(0.070000, 0.876000)
(0.080000, 0.871400)
(0.090000, 0.879700)
(0.100000, 0.866950)
(0.120000, 0.857000)
(0.140000, 0.858100)
(0.150000, 0.850800)
(0.160000, 0.855600)
(0.180000, 0.850050)
(0.200000, 0.834000)
(0.210000, 0.841400)
(0.240000, 0.827550)
(0.270000, 0.824400)
(0.280000, 0.806600)
(0.300000, 0.781900)
(0.320000, 0.784400)
(0.350000, 0.750900)
(0.360000, 0.756300)
(0.400000, 0.681200)
(0.420000, 0.670000)
(0.450000, 0.636100)
(0.480000, 0.579900)
(0.500000, 0.403400)
(0.540000, 0.484000)
(0.560000, 0.507700)
(0.600000, 0.201300)
(0.630000, 0.291100)
(0.700000, 0.082100)
(0.720000, 0.250800)
(0.800000, 0.034000)
(0.900000, 0.016400)
};
\addlegendentry{Client dominant}
\addplot[color=blue,dashed,ultra thick,mark=none] coordinates { 
(0.020000, 0.871100)
(0.030000, 0.866000)
(0.040000, 0.865400)
(0.050000, 0.853600)
(0.060000, 0.850700)
(0.070000, 0.813300)
(0.080000, 0.823400)
(0.090000, 0.766800)
(0.100000, 0.802450)
(0.120000, 0.833700)
(0.140000, 0.793700)
(0.150000, 0.820500)
(0.160000, 0.745800)
(0.180000, 0.770200)
(0.200000, 0.756250)
(0.210000, 0.777000)
(0.240000, 0.761700)
(0.270000, 0.666800)
(0.280000, 0.741500)
(0.300000, 0.557050)
(0.320000, 0.692600)
(0.350000, 0.694500)
(0.360000, 0.601200)
(0.400000, 0.538800)
(0.420000, 0.619400)
(0.450000, 0.474500)
(0.480000, 0.477500)
(0.500000, 0.367700)
(0.540000, 0.402200)
(0.560000, 0.376400)
(0.600000, 0.336500)
(0.630000, 0.320600)
(0.700000, 0.261800)
(0.720000, 0.333400)
(0.800000, 0.264900)
(0.900000, 0.033400)
};
\addlegendentry{Flipping dominant}
\end{axis}
\end{tikzpicture}

%% file: figures/dom_SVC.tex
\begin{tikzpicture}[scale=0.475]
\begin{axis}[ 
		   width=0.7\textwidth,
		   height=0.5\textwidth,
	 	   x tick label style={scale=1.25,
			scale=0.95,
   		 	/pgf/number format/.cd,
			/pgf/number format/1000 sep={},
   			fixed,
   			fixed zerofill,
    			precision=1
		   },
		   x label style={scale=1.5},
	 	   y tick label style={scale=1.25,
    		 	/pgf/number format/.cd,
   			fixed,
   			fixed zerofill,
    			precision=1
		    },
		   y label style={scale=1.5},
                    xmin=0.0,xmax=0.92,
                    ymin=0.0,ymax=1.02,
                    xtick={0.0,0.1,0.2,0.3,0.4,0.5,0.6,0.7,0.8,0.9},
                    ytick={0.1,0.2,0.3,0.4,0.5,0.6,0.7,0.8,0.9,1.0},
                    xlabel={Label flipping rate},
                    ylabel={Accuracy},
                    legend cell align=left,
                    legend pos=north east,
                    ] 
\addplot[color=red,ultra thick,mark=none] coordinates { 
(0.020000, 0.789100)
(0.030000, 0.776000)
(0.040000, 0.790300)
(0.050000, 0.796200)
(0.060000, 0.764350)
(0.070000, 0.760400)
(0.080000, 0.778800)
(0.090000, 0.785600)
(0.100000, 0.791800)
(0.120000, 0.770900)
(0.140000, 0.782100)
(0.150000, 0.775000)
(0.160000, 0.773400)
(0.180000, 0.782050)
(0.200000, 0.780250)
(0.210000, 0.756200)
(0.240000, 0.772950)
(0.270000, 0.734000)
(0.280000, 0.728100)
(0.300000, 0.749850)
(0.320000, 0.693000)
(0.350000, 0.695400)
(0.360000, 0.645100)
(0.400000, 0.609900)
(0.420000, 0.579400)
(0.450000, 0.523000)
(0.480000, 0.387000)
(0.500000, 0.347600)
(0.540000, 0.243300)
(0.560000, 0.195800)
(0.600000, 0.182000)
(0.630000, 0.102000)
(0.700000, 0.058200)
(0.720000, 0.046700)
(0.800000, 0.033500)
(0.900000, 0.025600)
};
\addlegendentry{Client dominant}
\addplot[color=blue,dashed,ultra thick,mark=none] coordinates { 
(0.020000, 0.790000)
(0.030000, 0.790700)
(0.040000, 0.790000)
(0.050000, 0.791100)
(0.060000, 0.787750)
(0.070000, 0.789500)
(0.080000, 0.790100)
(0.090000, 0.789500)
(0.100000, 0.789300)
(0.120000, 0.786100)
(0.140000, 0.789100)
(0.150000, 0.795300)
(0.160000, 0.786000)
(0.180000, 0.784800)
(0.200000, 0.781800)
(0.210000, 0.778000)
(0.240000, 0.780100)
(0.270000, 0.765200)
(0.280000, 0.763500)
(0.300000, 0.736500)
(0.320000, 0.723100)
(0.350000, 0.678900)
(0.360000, 0.687000)
(0.400000, 0.596000)
(0.420000, 0.598200)
(0.450000, 0.521300)
(0.480000, 0.362300)
(0.500000, 0.238700)
(0.540000, 0.267000)
(0.560000, 0.198400)
(0.600000, 0.125900)
(0.630000, 0.115400)
(0.700000, 0.062100)
(0.720000, 0.067800)
(0.800000, 0.033000)
(0.900000, 0.029300)
};
\addlegendentry{Flipping dominant}
\end{axis}
\end{tikzpicture}

%% file: figures/dom_MLP.tex
\begin{tikzpicture}[scale=0.475]
\begin{axis}[ 
		   width=0.7\textwidth,
		   height=0.5\textwidth,
	 	   x tick label style={scale=1.25,
			scale=0.95,
   		 	/pgf/number format/.cd,
			/pgf/number format/1000 sep={},
   			fixed,
   			fixed zerofill,
    			precision=1
		   },
		   x label style={scale=1.5},
	 	   y tick label style={scale=1.25,
    		 	/pgf/number format/.cd,
   			fixed,
   			fixed zerofill,
    			precision=1
		    },
		   y label style={scale=1.5},
                    xmin=0.0,xmax=0.92,
                    ymin=0.0,ymax=1.02,
                    xtick={0.0,0.1,0.2,0.3,0.4,0.5,0.6,0.7,0.8,0.9},
                    ytick={0.1,0.2,0.3,0.4,0.5,0.6,0.7,0.8,0.9,1.0},
                    xlabel={Label flipping rate},
                    ylabel={Accuracy},
                    legend cell align=left,
                    legend pos=north east,
                    ] 
\addplot[color=red,ultra thick,mark=none] coordinates { 
(0.020000, 0.921400)
(0.030000, 0.919000)
(0.040000, 0.907500)
(0.050000, 0.906900)
(0.060000, 0.911250)
(0.070000, 0.900900)
(0.080000, 0.904400)
(0.090000, 0.897900)
(0.100000, 0.892150)
(0.120000, 0.898700)
(0.140000, 0.851200)
(0.150000, 0.875500)
(0.160000, 0.835600)
(0.180000, 0.746700)
(0.200000, 0.597350)
(0.210000, 0.533600)
(0.240000, 0.267350)
(0.270000, 0.055800)
(0.280000, 0.133900)
(0.300000, 0.061700)
(0.320000, 0.034300)
(0.350000, 0.049000)
(0.360000, 0.017900)
(0.400000, 0.014000)
(0.420000, 0.040900)
(0.450000, 0.006000)
(0.480000, 0.013300)
(0.500000, 0.006300)
(0.540000, 0.005300)
(0.560000, 0.018700)
(0.600000, 0.004900)
(0.630000, 0.008500)
(0.700000, 0.006500)
(0.720000, 0.006700)
(0.800000, 0.006100)
(0.900000, 0.004700)
};
\addlegendentry{Client dominant}
\addplot[color=blue,dashed,ultra thick,mark=none] coordinates { 
(0.020000, 0.937300)
(0.030000, 0.938800)
(0.040000, 0.934000)
(0.050000, 0.938400)
(0.060000, 0.930350)
(0.070000, 0.932900)
(0.080000, 0.927900)
(0.090000, 0.933700)
(0.100000, 0.921900)
(0.120000, 0.918850)
(0.140000, 0.929800)
(0.150000, 0.903000)
(0.160000, 0.928500)
(0.180000, 0.908850)
(0.200000, 0.879300)
(0.210000, 0.890400)
(0.240000, 0.874700)
(0.270000, 0.898800)
(0.280000, 0.758800)
(0.300000, 0.646250)
(0.320000, 0.773500)
(0.350000, 0.488500)
(0.360000, 0.777700)
(0.400000, 0.702500)
(0.420000, 0.132500)
(0.450000, 0.570500)
(0.480000, 0.101500)
(0.500000, 0.604600)
(0.540000, 0.254000)
(0.560000, 0.030900)
(0.600000, 0.238400)
(0.630000, 0.032700)
(0.700000, 0.034600)
(0.720000, 0.014000)
(0.800000, 0.014700)
(0.900000, 0.006700)
};
\addlegendentry{Flipping dominant}
\end{axis}
\end{tikzpicture}

%% file: figures/dom_CNN.tex
\begin{tikzpicture}[scale=0.475]
\begin{axis}[ 
		   width=0.7\textwidth,
		   height=0.5\textwidth,
	 	   x tick label style={scale=1.25,
			scale=0.95,
   		 	/pgf/number format/.cd,
			/pgf/number format/1000 sep={},
   			fixed,
   			fixed zerofill,
    			precision=1
		   },
		   x label style={scale=1.5},
	 	   y tick label style={scale=1.25,
    		 	/pgf/number format/.cd,
   			fixed,
   			fixed zerofill,
    			precision=1
		    },
		   y label style={scale=1.5},
                    xmin=0.0,xmax=0.92,
                    ymin=0.0,ymax=1.02,
                    xtick={0.0,0.1,0.2,0.3,0.4,0.5,0.6,0.7,0.8,0.9},
                    ytick={0.1,0.2,0.3,0.4,0.5,0.6,0.7,0.8,0.9,1.0},
                    xlabel={Label flipping rate},
                    ylabel={Accuracy},
                    legend cell align=left,
                    legend pos=north east,
                    ] 
\addplot[color=red,ultra thick,mark=none] coordinates { 
(0.020000, 0.891100)
(0.030000, 0.744600)
(0.040000, 0.522500)
(0.050000, 0.733600)
(0.060000, 0.789550)
(0.070000, 0.874800)
(0.080000, 0.677450)
(0.090000, 0.647800)
(0.100000, 0.527800)
(0.120000, 0.708000)
(0.140000, 0.725100)
(0.150000, 0.747400)
(0.160000, 0.599100)
(0.180000, 0.548100)
(0.200000, 0.530400)
(0.210000, 0.498000)
(0.240000, 0.494500)
(0.270000, 0.398700)
(0.280000, 0.453700)
(0.300000, 0.241550)
(0.320000, 0.330900)
(0.350000, 0.196300)
(0.360000, 0.209300)
(0.400000, 0.149000)
(0.420000, 0.181000)
(0.450000, 0.093700)
(0.480000, 0.060800)
(0.500000, 0.103000)
(0.540000, 0.083200)
(0.560000, 0.148600)
(0.600000, 0.077500)
(0.630000, 0.079500)
(0.700000, 0.103300)
(0.720000, 0.038900)
(0.800000, 0.058400)
(0.900000, 0.099500)
};
\addlegendentry{Client dominant}
\addplot[color=blue,dashed,ultra thick,mark=none] coordinates { 
(0.020000, 0.926900)
(0.030000, 0.930900)
(0.040000, 0.934100)
(0.050000, 0.923200)
(0.060000, 0.922050)
(0.070000, 0.939800)
(0.080000, 0.882400)
(0.090000, 0.930900)
(0.100000, 0.887450)
(0.120000, 0.813550)
(0.140000, 0.865000)
(0.150000, 0.556900)
(0.160000, 0.900700)
(0.180000, 0.723150)
(0.200000, 0.633950)
(0.210000, 0.576300)
(0.240000, 0.377450)
(0.270000, 0.510700)
(0.280000, 0.457800)
(0.300000, 0.492250)
(0.320000, 0.340400)
(0.350000, 0.502700)
(0.360000, 0.338200)
(0.400000, 0.413500)
(0.420000, 0.445800)
(0.450000, 0.421500)
(0.480000, 0.319200)
(0.500000, 0.553400)
(0.540000, 0.366500)
(0.560000, 0.151600)
(0.600000, 0.363600)
(0.630000, 0.230000)
(0.700000, 0.223400)
(0.720000, 0.112500)
(0.800000, 0.118000)
(0.900000, 0.108100)
};
\addlegendentry{Flipping dominant}
\end{axis}
\end{tikzpicture}

%% file: figures/dom_RF.tex
\begin{tikzpicture}[scale=0.475]
\begin{axis}[ 
		   width=0.7\textwidth,
		   height=0.5\textwidth,
	 	   x tick label style={scale=1.25,
			scale=0.95,
   		 	/pgf/number format/.cd,
			/pgf/number format/1000 sep={},
   			fixed,
   			fixed zerofill,
    			precision=1
		   },
		   x label style={scale=1.5},
	 	   y tick label style={scale=1.25,
    		 	/pgf/number format/.cd,
   			fixed,
   			fixed zerofill,
    			precision=1
		    },
		   y label style={scale=1.5},
                    xmin=0.0,xmax=0.92,
                    ymin=0.0,ymax=1.02,
                    xtick={0.0,0.1,0.2,0.3,0.4,0.5,0.6,0.7,0.8,0.9},
                    ytick={0.1,0.2,0.3,0.4,0.5,0.6,0.7,0.8,0.9,1.0},
                    xlabel={Label flipping rate},
                    ylabel={Accuracy},
                    legend cell align=left,
                    legend pos=north east,
                    ] 
\addplot[color=red,ultra thick,mark=none] coordinates { 
(0.020000, 0.849100)
(0.030000, 0.849500)
(0.040000, 0.849100)
(0.050000, 0.832600)
(0.060000, 0.830550)
(0.070000, 0.845900)
(0.080000, 0.847350)
(0.090000, 0.835900)
(0.100000, 0.832200)
(0.120000, 0.819250)
(0.140000, 0.834100)
(0.150000, 0.812000)
(0.160000, 0.804200)
(0.180000, 0.760200)
(0.200000, 0.786650)
(0.210000, 0.798100)
(0.240000, 0.655950)
(0.270000, 0.713200)
(0.280000, 0.618100)
(0.300000, 0.572350)
(0.320000, 0.600800)
(0.350000, 0.431800)
(0.360000, 0.542500)
(0.400000, 0.532750)
(0.420000, 0.267400)
(0.450000, 0.397500)
(0.480000, 0.261500)
(0.500000, 0.423500)
(0.540000, 0.258000)
(0.560000, 0.172200)
(0.600000, 0.256800)
(0.630000, 0.160200)
(0.700000, 0.153200)
(0.720000, 0.125700)
(0.800000, 0.108000)
(0.900000, 0.031000)
};
\addlegendentry{Client dominant}
\addplot[color=blue,dashed,ultra thick,mark=none] coordinates { 
(0.020000, 0.853700)
(0.030000, 0.843200)
(0.040000, 0.855000)
(0.050000, 0.853900)
(0.060000, 0.853550)
(0.070000, 0.845100)
(0.080000, 0.843450)
(0.090000, 0.851900)
(0.100000, 0.838300)
(0.120000, 0.821900)
(0.140000, 0.809400)
(0.150000, 0.811100)
(0.160000, 0.798700)
(0.180000, 0.797450)
(0.200000, 0.794800)
(0.210000, 0.793500)
(0.240000, 0.785600)
(0.270000, 0.776300)
(0.280000, 0.777100)
(0.300000, 0.191400)
(0.320000, 0.773700)
(0.350000, 0.767800)
(0.360000, 0.117300)
(0.400000, 0.124550)
(0.420000, 0.177000)
(0.450000, 0.116700)
(0.480000, 0.135200)
(0.500000, 0.108300)
(0.540000, 0.114500)
(0.560000, 0.130600)
(0.600000, 0.103700)
(0.630000, 0.112200)
(0.700000, 0.102900)
(0.720000, 0.110800)
(0.800000, 0.094700)
(0.900000, 0.028300)
};
\addlegendentry{Flipping dominant}
\end{axis}
\end{tikzpicture}

%% file: figures/dom_XGB.tex
\begin{tikzpicture}[scale=0.475]
\begin{axis}[ 
		   width=0.7\textwidth,
		   height=0.5\textwidth,
	 	   x tick label style={scale=1.25,
			scale=0.95,
   		 	/pgf/number format/.cd,
			/pgf/number format/1000 sep={},
   			fixed,
   			fixed zerofill,
    			precision=1
		   },
		   x label style={scale=1.5},
	 	   y tick label style={scale=1.25,
    		 	/pgf/number format/.cd,
   			fixed,
   			fixed zerofill,
    			precision=1
		    },
		   y label style={scale=1.5},
                    xmin=0.0,xmax=0.92,
                    ymin=0.0,ymax=1.02,
                    xtick={0.0,0.1,0.2,0.3,0.4,0.5,0.6,0.7,0.8,0.9},
                    ytick={0.1,0.2,0.3,0.4,0.5,0.6,0.7,0.8,0.9,1.0},
                    xlabel={Label flipping rate},
                    ylabel={Accuracy},
                    legend cell align=left,
                    legend pos=north east,
                    ] 
\addplot[color=red,ultra thick,mark=none] coordinates { 
(0.020000, 0.817600)
(0.030000, 0.817200)
(0.040000, 0.816100)
(0.050000, 0.810500)
(0.060000, 0.805750)
(0.070000, 0.812300)
(0.080000, 0.803800)
(0.090000, 0.814800)
(0.100000, 0.785050)
(0.120000, 0.796650)
(0.140000, 0.794900)
(0.150000, 0.770500)
(0.160000, 0.739900)
(0.180000, 0.726950)
(0.200000, 0.756450)
(0.210000, 0.682200)
(0.240000, 0.641000)
(0.270000, 0.679000)
(0.280000, 0.597300)
(0.300000, 0.565900)
(0.320000, 0.615100)
(0.350000, 0.445000)
(0.360000, 0.601300)
(0.400000, 0.503350)
(0.420000, 0.249800)
(0.450000, 0.465700)
(0.480000, 0.273800)
(0.500000, 0.390900)
(0.540000, 0.252700)
(0.560000, 0.194600)
(0.600000, 0.230800)
(0.630000, 0.173000)
(0.700000, 0.176000)
(0.720000, 0.132800)
(0.800000, 0.100800)
(0.900000, 0.043100)
};
\addlegendentry{Client dominant}
\addplot[color=blue,dashed,ultra thick,mark=none] coordinates { 
(0.020000, 0.828200)
(0.030000, 0.831400)
(0.040000, 0.816800)
(0.050000, 0.830200)
(0.060000, 0.815200)
(0.070000, 0.822800)
(0.080000, 0.813750)
(0.090000, 0.814700)
(0.100000, 0.811150)
(0.120000, 0.785950)
(0.140000, 0.786800)
(0.150000, 0.779200)
(0.160000, 0.774600)
(0.180000, 0.773150)
(0.200000, 0.775100)
(0.210000, 0.762400)
(0.240000, 0.760800)
(0.270000, 0.755700)
(0.280000, 0.758700)
(0.300000, 0.658200)
(0.320000, 0.649500)
(0.350000, 0.322800)
(0.360000, 0.528200)
(0.400000, 0.236100)
(0.420000, 0.187600)
(0.450000, 0.136700)
(0.480000, 0.149200)
(0.500000, 0.128900)
(0.540000, 0.139800)
(0.560000, 0.147800)
(0.600000, 0.137500)
(0.630000, 0.123300)
(0.700000, 0.102800)
(0.720000, 0.118100)
(0.800000, 0.085800)
(0.900000, 0.024200)
};
\addlegendentry{Flipping dominant}
\end{axis}
\end{tikzpicture}

%% file: figures/dom_LSTM.tex
\begin{tikzpicture}[scale=0.475]
\begin{axis}[ 
		   width=0.7\textwidth,
		   height=0.5\textwidth,
	 	   x tick label style={scale=1.25,
			scale=0.95,
   		 	/pgf/number format/.cd,
			/pgf/number format/1000 sep={},
   			fixed,
   			fixed zerofill,
    			precision=1
		   },
		   x label style={scale=1.5},
	 	   y tick label style={scale=1.25,
    		 	/pgf/number format/.cd,
   			fixed,
   			fixed zerofill,
    			precision=1
		    },
		   y label style={scale=1.5},
                    xmin=0.0,xmax=0.92,
                    ymin=0.0,ymax=1.02,
                    xtick={0.0,0.1,0.2,0.3,0.4,0.5,0.6,0.7,0.8,0.9},
                    ytick={0.1,0.2,0.3,0.4,0.5,0.6,0.7,0.8,0.9,1.0},
                    xlabel={Label flipping rate},
                    ylabel={Accuracy},
                    legend cell align=left,
                    legend pos=north east,
                    ] 
\addplot[color=red,ultra thick,mark=none] coordinates { 
(0.070000, 0.943600)
(0.080000, 0.943400)
(0.090000, 0.939000)
(0.100000, 0.938450)
(0.120000, 0.940450)
(0.140000, 0.935000)
(0.150000, 0.939300)
(0.160000, 0.934600)
(0.180000, 0.932700)
(0.200000, 0.935350)
(0.210000, 0.930500)
(0.240000, 0.925000)
(0.270000, 0.918000)
(0.280000, 0.883300)
(0.300000, 0.633000)
(0.320000, 0.754400)
(0.350000, 0.143200)
(0.360000, 0.842300)
(0.400000, 0.385000)
(0.420000, 0.062500)
(0.450000, 0.012900)
(0.480000, 0.016300)
(0.500000, 0.004900)
(0.540000, 0.006100)
(0.560000, 0.012700)
(0.600000, 0.005200)
(0.630000, 0.006600)
(0.700000, 0.003700)
(0.720000, 0.008700)
(0.800000, 0.004500)
(0.900000, 0.004600)
};
\addlegendentry{Client dominant}
\addplot[color=blue,dashed,ultra thick,mark=none] coordinates { 
(0.020000, 0.948800)
(0.030000, 0.948900)
(0.040000, 0.948600)
(0.050000, 0.942300)
(0.060000, 0.944850)
(0.070000, 0.946900)
(0.080000, 0.944600)
(0.090000, 0.947800)
(0.100000, 0.941900)
(0.120000, 0.938600)
(0.140000, 0.944000)
(0.150000, 0.916600)
(0.160000, 0.934800)
(0.180000, 0.914800)
(0.200000, 0.890100)
(0.210000, 0.912200)
(0.240000, 0.846950)
(0.270000, 0.870100)
(0.280000, 0.671600)
(0.300000, 0.649650)
(0.320000, 0.725400)
(0.350000, 0.440600)
(0.360000, 0.753000)
(0.400000, 0.623100)
(0.420000, 0.219200)
(0.450000, 0.519800)
(0.480000, 0.265900)
(0.500000, 0.471800)
(0.540000, 0.146400)
(0.560000, 0.040700)
(0.600000, 0.192500)
(0.630000, 0.029000)
(0.700000, 0.050100)
(0.720000, 0.009300)
(0.800000, 0.013300)
(0.900000, 0.006500)
};
\addlegendentry{Flipping dominant}
\end{axis}
\end{tikzpicture}

%% file: figures/3d_10_MLR.tex
\begin{tikzpicture}[xscale=1.0,yscale=1.0]
\begin{axis}[%
view={10}{26},
width=10cm,
height=6cm,
scale only axis,
xmin=1, xmax=10,
xmajorgrids,
ymin=10, ymax=100,
ymajorgrids,
zmin=0.0, zmax=1.0,
zmajorgrids,
axis lines=left,
grid=major,
xlabel=Adversarial clients,
xlabel style={yshift=0.2cm},
ylabel=Percent labels flipped,
ylabel style={yshift=0.5cm,xshift=0.15cm,rotate=55},
xtick={10, 9, 8, 7, 6, 5, 4, 3, 2, 1},
ytick={100, 70, 40, 10},
x dir=reverse,
y dir=reverse,
zlabel=Accuracy,
zlabel style={xshift=0.65cm,yshift=0.0cm},
x tick label style={xshift=0.05cm,scale=0.9,
    		/pgf/number format/.cd,
		1000 sep={},
    		fixed,
    		fixed zerofill,
    		precision=0},
y tick label style={yshift=0.25cm,xshift=0.075cm,scale=0.9,
    		/pgf/number format/.cd,
		1000 sep={},
    		fixed,
    		fixed zerofill,
    		precision=0},
z tick label style={yshift=0.4cm,scale=0.9,
    		/pgf/number format/.cd,
		1000 sep={},
    		fixed,
    		fixed zerofill,
    		precision=2}]
%
\addplot3[%
surf,
z buffer=sort,
colormap/jet,
shader=flat,
draw=black]
coordinates{ 

(10, 100, 0.0114)
(10, 90, 0.0164)
(10, 80, 0.034)
(10, 70, 0.0821)
(10, 60, 0.2013)
(10, 50, 0.4034)
(10, 40, 0.6477)
(10, 30, 0.785)
(10, 20, 0.8439)
(10, 10, 0.8777)

(9, 100, 0.0334)
(9, 90, 0.1415)
(9, 80, 0.2508)
(9, 70, 0.2911)
(9, 60, 0.484)
(9, 50, 0.6361)
(9, 40, 0.7563)
(9, 30, 0.8244)
(9, 20, 0.8537)
(9, 10, 0.8797)

(8, 100, 0.2649)
(8, 90, 0.3334)
(8, 80, 0.307)
(8, 70, 0.5077)
(8, 60, 0.5799)
(8, 50, 0.7147)
(8, 40, 0.7844)
(8, 30, 0.8352)
(8, 20, 0.8556)
(8, 10, 0.8768)

(7, 100, 0.2618)
(7, 90, 0.3206)
(7, 80, 0.3764)
(7, 70, 0.5208)
(7, 60, 0.67)
(7, 50, 0.7509)
(7, 40, 0.8066)
(7, 30, 0.8414)
(7, 20, 0.8581)
(7, 10, 0.876)

(6, 100, 0.3365)
(6, 90, 0.4022)
(6, 80, 0.4775)
(6, 70, 0.6194)
(6, 60, 0.7257)
(6, 50, 0.7788)
(6, 40, 0.8199)
(6, 30, 0.8464)
(6, 20, 0.8567)
(6, 10, 0.8791)

(5, 100, 0.3677)
(5, 90, 0.4745)
(5, 80, 0.5972)
(5, 70, 0.6945)
(5, 60, 0.7597)
(5, 50, 0.8037)
(5, 40, 0.8241)
(5, 30, 0.8508)
(5, 20, 0.8562)
(5, 10, 0.8796)

(4, 100, 0.4804)
(4, 90, 0.6012)
(4, 80, 0.6926)
(4, 70, 0.7415)
(4, 60, 0.7822)
(4, 50, 0.8064)
(4, 40, 0.8327)
(4, 30, 0.8573)
(4, 20, 0.866)
(4, 10, 0.8751)

(3, 100, 0.3544)
(3, 90, 0.6668)
(3, 80, 0.7412)
(3, 70, 0.777)
(3, 60, 0.8044)
(3, 50, 0.8205)
(3, 40, 0.8446)
(3, 30, 0.8521)
(3, 20, 0.8628)
(3, 10, 0.8755)

(2, 100, 0.7061)
(2, 90, 0.736)
(2, 80, 0.7458)
(2, 70, 0.7937)
(2, 60, 0.8228)
(2, 50, 0.8287)
(2, 40, 0.8455)
(2, 30, 0.8561)
(2, 20, 0.8674)
(2, 10, 0.8736)

(1, 100, 0.7762)
(1, 90, 0.7668)
(1, 80, 0.8013)
(1, 70, 0.8133)
(1, 60, 0.8453)
(1, 50, 0.8536)
(1, 40, 0.8654)
(1, 30, 0.866)
(1, 20, 0.8711)
(1, 10, 0.8726)

};
  \end{axis}
  \end{tikzpicture}

%% file: figures/3d_10_SVC.tex
\begin{tikzpicture}[xscale=1.0,yscale=1.0]
\begin{axis}[%
view={10}{26},
width=10cm,
height=6cm,
scale only axis,
xmin=1, xmax=10,
xmajorgrids,
ymin=10, ymax=100,
ymajorgrids,
zmin=0.0, zmax=1.0,
zmajorgrids,
axis lines=left,
grid=major,
xlabel=Adversarial clients,
xlabel style={yshift=0.2cm},
ylabel=Percent labels flipped,
ylabel style={yshift=0.5cm,xshift=0.15cm,rotate=55},
xtick={10, 9, 8, 7, 6, 5, 4, 3, 2, 1},
ytick={100, 70, 40, 10},
x dir=reverse,
y dir=reverse,
zlabel=Accuracy,
zlabel style={xshift=0.65cm,yshift=0.0cm},
x tick label style={xshift=0.05cm,scale=0.9,
    		/pgf/number format/.cd,
		1000 sep={},
    		fixed,
    		fixed zerofill,
    		precision=0},
y tick label style={yshift=0.25cm,xshift=0.075cm,scale=0.9,
    		/pgf/number format/.cd,
		1000 sep={},
    		fixed,
    		fixed zerofill,
    		precision=0},
z tick label style={yshift=0.4cm,scale=0.9,
    		/pgf/number format/.cd,
		1000 sep={},
    		fixed,
    		fixed zerofill,
    		precision=2}]
%


\addplot3[%
surf,
z buffer=sort,
colormap/jet,
shader=flat,
draw=black]
coordinates{ 

(10, 100, 0.0214)
(10, 90, 0.0256)
(10, 80, 0.0335)
(10, 70, 0.0582)
(10, 60, 0.182)
(10, 50, 0.3476)
(10, 40, 0.6056)
(10, 30, 0.7214)
(10, 20, 0.783)
(10, 10, 0.7922)

(9, 100, 0.0293)
(9, 90, 0.0324)
(9, 80, 0.0467)
(9, 70, 0.102)
(9, 60, 0.2433)
(9, 50, 0.523)
(9, 40, 0.6451)
(9, 30, 0.734)
(9, 20, 0.7841)
(9, 10, 0.7856)

(8, 100, 0.033)
(8, 90, 0.0678)
(8, 80, 0.098)
(8, 70, 0.1958)
(8, 60, 0.387)
(8, 50, 0.6142)
(8, 40, 0.693)
(8, 30, 0.7605)
(8, 20, 0.7734)
(8, 10, 0.7922)

(7, 100, 0.0621)
(7, 90, 0.1154)
(7, 80, 0.1984)
(7, 70, 0.3126)
(7, 60, 0.5794)
(7, 50, 0.6954)
(7, 40, 0.7281)
(7, 30, 0.7562)
(7, 20, 0.7821)
(7, 10, 0.7604)

(6, 100, 0.1259)
(6, 90, 0.267)
(6, 80, 0.3623)
(6, 70, 0.5982)
(6, 60, 0.6791)
(6, 50, 0.7783)
(6, 40, 0.7854)
(6, 30, 0.78)
(6, 20, 0.756)
(6, 10, 0.7401)

(5, 100, 0.2387)
(5, 90, 0.5213)
(5, 80, 0.662)
(5, 70, 0.6789)
(5, 60, 0.7476)
(5, 50, 0.7865)
(5, 40, 0.7775)
(5, 30, 0.775)
(5, 20, 0.7914)
(5, 10, 0.7962)

(4, 100, 0.53)
(4, 90, 0.687)
(4, 80, 0.7231)
(4, 70, 0.7635)
(4, 60, 0.7829)
(4, 50, 0.7783)
(4, 40, 0.7903)
(4, 30, 0.7858)
(4, 20, 0.7654)
(4, 10, 0.7903)

(3, 100, 0.7254)
(3, 90, 0.7652)
(3, 80, 0.7773)
(3, 70, 0.778)
(3, 60, 0.7862)
(3, 50, 0.7953)
(3, 40, 0.7869)
(3, 30, 0.7952)
(3, 20, 0.7886)
(3, 10, 0.776)

(2, 100, 0.7853)
(2, 90, 0.7834)
(2, 80, 0.786)
(2, 70, 0.7891)
(2, 60, 0.7853)
(2, 50, 0.7877)
(2, 40, 0.7902)
(2, 30, 0.786)
(2, 20, 0.7832)
(2, 10, 0.7891)

(1, 100, 0.7909)
(1, 90, 0.7895)
(1, 80, 0.79)
(1, 70, 0.7895)
(1, 60, 0.7895)
(1, 50, 0.7911)
(1, 40, 0.79)
(1, 30, 0.7907)
(1, 20, 0.79)
(1, 10, 0.7897)

};
  \end{axis}
  \end{tikzpicture}

%% file: figures/3d_10_MLP.tex
\begin{tikzpicture}[xscale=1.0,yscale=1.0]
\begin{axis}[%
view={10}{26},
width=10cm,
height=6cm,
scale only axis,
xmin=1, xmax=10,
xmajorgrids,
ymin=10, ymax=100,
ymajorgrids,
zmin=0.0, zmax=1.0,
zmajorgrids,
axis lines=left,
grid=major,
xlabel=Adversarial clients,
xlabel style={yshift=0.2cm},
ylabel=Percent labels flipped,
ylabel style={yshift=0.5cm,xshift=0.15cm,rotate=55},
xtick={10, 9, 8, 7, 6, 5, 4, 3, 2, 1},
ytick={100, 70, 40, 10},
x dir=reverse,
y dir=reverse,
zlabel=Accuracy,
zlabel style={xshift=0.65cm,yshift=0.0cm},
x tick label style={xshift=0.05cm,scale=0.9,
    		/pgf/number format/.cd,
		1000 sep={},
    		fixed,
    		fixed zerofill,
    		precision=0},
y tick label style={yshift=0.25cm,xshift=0.075cm,scale=0.9,
    		/pgf/number format/.cd,
		1000 sep={},
    		fixed,
    		fixed zerofill,
    		precision=0},
z tick label style={yshift=0.4cm,scale=0.9,
    		/pgf/number format/.cd,
		1000 sep={},
    		fixed,
    		fixed zerofill,
    		precision=2}]
%
\addplot3[%
surf,
z buffer=sort,
colormap/jet,
shader=flat,
draw=black]
coordinates{ 

(10, 100, 0.0029)
(10, 90, 0.0035)
(10, 80, 0.0026)
(10, 70, 0.0023)
(10, 60, 0.0032)
(10, 50, 0.0027)
(10, 40, 0.9098)
(10, 30, 0.9312)
(10, 20, 0.924)
(10, 10, 0.948)

(9, 100, 0.0049)
(9, 90, 0.0042)
(9, 80, 0.0043)
(9, 70, 0.004)
(9, 60, 0.0048)
(9, 50, 0.0066)
(9, 40, 0.8814)
(9, 30, 0.9309)
(9, 20, 0.9348)
(9, 10, 0.9463)

(8, 100, 0.0074)
(8, 90, 0.0087)
(8, 80, 0.0074)
(8, 70, 0.0057)
(8, 60, 0.0068)
(8, 50, 0.0059)
(8, 40, 0.9153)
(8, 30, 0.9334)
(8, 20, 0.9464)
(8, 10, 0.9512)

(7, 100, 0.016)
(7, 90, 0.0132)
(7, 80, 0.016)
(7, 70, 0.0159)
(7, 60, 0.0185)
(7, 50, 0.0223)
(7, 40, 0.906)
(7, 30, 0.9378)
(7, 20, 0.9451)
(7, 10, 0.9457)

(6, 100, 0.0518)
(6, 90, 0.0546)
(6, 80, 0.0922)
(6, 70, 0.0867)
(6, 60, 0.0812)
(6, 50, 0.0717)
(6, 40, 0.9304)
(6, 30, 0.9486)
(6, 20, 0.9507)
(6, 10, 0.9519)

(5, 100, 0.5776)
(5, 90, 0.5651)
(5, 80, 0.5403)
(5, 70, 0.609)
(5, 60, 0.5932)
(5, 50, 0.5662)
(5, 40, 0.9336)
(5, 30, 0.9463)
(5, 20, 0.9465)
(5, 10, 0.9619)

(4, 100, 0.9053)
(4, 90, 0.9325)
(4, 80, 0.9186)
(4, 70, 0.9226)
(4, 60, 0.9136)
(4, 50, 0.9316)
(4, 40, 0.944)
(4, 30, 0.956)
(4, 20, 0.9555)
(4, 10, 0.9634)

(3, 100, 0.952)
(3, 90, 0.9547)
(3, 80, 0.9488)
(3, 70, 0.9532)
(3, 60, 0.9436)
(3, 50, 0.9561)
(3, 40, 0.9583)
(3, 30, 0.954)
(3, 20, 0.9635)
(3, 10, 0.9638)

(2, 100, 0.9613)
(2, 90, 0.9598)
(2, 80, 0.9635)
(2, 70, 0.9564)
(2, 60, 0.966)
(2, 50, 0.9663)
(2, 40, 0.9608)
(2, 30, 0.9596)
(2, 20, 0.9635)
(2, 10, 0.9639)

(1, 100, 0.9674)
(1, 90, 0.9706)
(1, 80, 0.9636)
(1, 70, 0.9677)
(1, 60, 0.9659)
(1, 50, 0.9684)
(1, 40, 0.9665)
(1, 30, 0.9679)
(1, 20, 0.971)
(1, 10, 0.9666)

};
  \end{axis}
  \end{tikzpicture}

%% file: figures/3d_10_CNN.tex
\begin{tikzpicture}[xscale=1.0,yscale=1.0]
\begin{axis}[%
view={10}{26},
width=10cm,
height=6cm,
scale only axis,
xmin=1, xmax=10,
xmajorgrids,
ymin=10, ymax=100,
ymajorgrids,
zmin=0.0, zmax=1.0,
zmajorgrids,
axis lines=left,
grid=major,
xlabel=Adversarial clients,
xlabel style={yshift=0.2cm},
ylabel=Percent labels flipped,
ylabel style={yshift=0.5cm,xshift=0.15cm,rotate=55},
xtick={10, 9, 8, 7, 6, 5, 4, 3, 2, 1},
ytick={100, 70, 40, 10},
x dir=reverse,
y dir=reverse,
zlabel=Accuracy,
zlabel style={xshift=0.65cm,yshift=0.0cm},
x tick label style={xshift=0.05cm,scale=0.9,
    		/pgf/number format/.cd,
		1000 sep={},
    		fixed,
    		fixed zerofill,
    		precision=0},
y tick label style={yshift=0.25cm,xshift=0.075cm,scale=0.9,
    		/pgf/number format/.cd,
		1000 sep={},
    		fixed,
    		fixed zerofill,
    		precision=0},
z tick label style={yshift=0.4cm,scale=0.9,
    		/pgf/number format/.cd,
		1000 sep={},
    		fixed,
    		fixed zerofill,
    		precision=2}]
%
\addplot3[%
surf,
z buffer=sort,
colormap/jet,
shader=flat,
draw=black]
coordinates{ 

(10, 100, 0.0023)
(10, 90, 0.0021)
(10, 80, 0.0029)
(10, 70, 0.0027)
(10, 60, 0.0025)
(10, 50, 0.0045)
(10, 40, 0.8442)
(10, 30, 0.9162)
(10, 20, 0.9442)
(10, 10, 0.9438)

(9, 100, 0.0057)
(9, 90, 0.0066)
(9, 80, 0.004)
(9, 70, 0.0054)
(9, 60, 0.006)
(9, 50, 0.0046)
(9, 40, 0.8591)
(9, 30, 0.9311)
(9, 20, 0.9474)
(9, 10, 0.9423)

(8, 100, 0.0127)
(8, 90, 0.014)
(8, 80, 0.0202)
(8, 70, 0.036)
(8, 60, 0.0123)
(8, 50, 0.0131)
(8, 40, 0.8227)
(8, 30, 0.9356)
(8, 20, 0.9061)
(8, 10, 0.9507)

(7, 100, 0.1427)
(7, 90, 0.0883)
(7, 80, 0.0721)
(7, 70, 0.0975)
(7, 60, 0.097)
(7, 50, 0.0847)
(7, 40, 0.8348)
(7, 30, 0.9295)
(7, 20, 0.9494)
(7, 10, 0.9466)

(6, 100, 0.3175)
(6, 90, 0.3626)
(6, 80, 0.3325)
(6, 70, 0.3636)
(6, 60, 0.3118)
(6, 50, 0.2922)
(6, 40, 0.8991)
(6, 30, 0.9286)
(6, 20, 0.9428)
(6, 10, 0.9387)

(5, 100, 0.4161)
(5, 90, 0.4355)
(5, 80, 0.3626)
(5, 70, 0.4948)
(5, 60, 0.4157)
(5, 50, 0.3974)
(5, 40, 0.8675)
(5, 30, 0.9081)
(5, 20, 0.9187)
(5, 10, 0.9394)

(4, 100, 0.6597)
(4, 90, 0.5311)
(4, 80, 0.923)
(4, 70, 0.7799)
(4, 60, 0.5265)
(4, 50, 0.6834)
(4, 40, 0.9728)
(4, 30, 0.9616)
(4, 20, 0.96)
(4, 10, 0.9342)

(3, 100, 0.8875)
(3, 90, 0.7885)
(3, 80, 0.8414)
(3, 70, 0.7312)
(3, 60, 0.7331)
(3, 50, 0.8108)
(3, 40, 0.923)
(3, 30, 0.8441)
(3, 20, 0.9845)
(3, 10, 0.9028)

(2, 100, 0.9429)
(2, 90, 0.9526)
(2, 80, 0.9498)
(2, 70, 0.922)
(2, 60, 0.9315)
(2, 50, 0.933)
(2, 40, 0.9686)
(2, 30, 0.9316)
(2, 20, 0.9667)
(2, 10, 0.955)

(1, 100, 0.9862)
(1, 90, 0.986)
(1, 80, 0.984)
(1, 70, 0.9844)
(1, 60, 0.9852)
(1, 50, 0.9839)
(1, 40, 0.9854)
(1, 30, 0.9868)
(1, 20, 0.9864)
(1, 10, 0.9842)

};
  \end{axis}
  \end{tikzpicture}

%% file: figures/3d_10_RF.tex
\begin{tikzpicture}[xscale=1.0,yscale=1.0]
\begin{axis}[%
view={10}{26},
width=10cm,
height=6cm,
scale only axis,
xmin=1, xmax=10,
xmajorgrids,
ymin=10, ymax=100,
ymajorgrids,
zmin=0.1, zmax=1,
zmajorgrids,
axis lines=left,
grid=major,
xlabel=Adversarial clients,
xlabel style={yshift=0.2cm},
ylabel=Percent labels flipped,
ylabel style={yshift=0.5cm,xshift=0.15cm,rotate=55},
xtick={10, 9, 8, 7, 6, 5, 4, 3, 2, 1},
ytick={100, 70, 40, 10},
x dir=reverse,
y dir=reverse,
zlabel=Accuracy,
zlabel style={xshift=0.65cm,yshift=0.0cm},
x tick label style={xshift=0.05cm,scale=0.9,
    		/pgf/number format/.cd,
		1000 sep={},
    		fixed,
    		fixed zerofill,
    		precision=0},
y tick label style={yshift=0.25cm,xshift=0.075cm,scale=0.9,
    		/pgf/number format/.cd,
		1000 sep={},
    		fixed,
    		fixed zerofill,
    		precision=0},
z tick label style={yshift=0.4cm,scale=0.9,
    		/pgf/number format/.cd,
		1000 sep={},
    		fixed,
    		fixed zerofill,
    		precision=2}]
%
\addplot3[%
surf,
z buffer=sort,
colormap/jet,
shader=flat,
draw=black]
coordinates{ 

(1, 10, 0.8489)
(1, 20, 0.8537)
(1, 30, 0.8432)
(1, 40, 0.8550)
(1, 50, 0.8539)
(1, 60, 0.8528)
(1, 70, 0.8451)
(1, 80, 0.8443)
(1, 90, 0.8519)
(1, 100, 0.8397)

(2, 10, 0.8491)
(2, 20, 0.8438)
(2, 30, 0.8543)
(2, 40, 0.8426)
(2, 50, 0.8369)
(2, 60, 0.8156)
(2, 70, 0.8094)
(2, 80, 0.7987)
(2, 90, 0.7930)
(2, 100, 0.7914)

(3, 10, 0.8495)
(3, 20, 0.8282)
(3, 30, 0.8455)
(3, 40, 0.8282)
(3, 50, 0.8111)
(3, 60, 0.8019)
(3, 70, 0.7935)
(3, 80, 0.7817)
(3, 90, 0.7763)
(3, 100, 0.1110)

(4, 10, 0.8491)
(4, 20, 0.8470)
(4, 30, 0.8334)
(4, 40, 0.8128)
(4, 50, 0.7982)
(4, 60, 0.7895)
(4, 70, 0.7771)
(4, 80, 0.7737)
(4, 90, 0.1173)
(4, 100, 0.1093)

(5, 10, 0.8326)
(5, 20, 0.8135)
(5, 30, 0.8120)
(5, 40, 0.7936)
(5, 50, 0.7842)
(5, 60, 0.2718)
(5, 70, 0.7678)
(5, 80, 0.1398)
(5, 90, 0.1167)
(5, 100, 0.1083)

(6, 10, 0.8329)
(6, 20, 0.8051)
(6, 30, 0.7272)
(6, 40, 0.6080)
(6, 50, 0.4351)
(6, 60, 0.2692)
(6, 70, 0.177)
(6, 80, 0.1352)
(6, 90, 0.1145)
(6, 100, 0.1037)

(7, 10, 0.8459)
(7, 20, 0.8341)
(7, 30, 0.7981)
(7, 40, 0.6181)
(7, 50, 0.4318)
(7, 60, 0.2674)
(7, 70, 0.1639)
(7, 80, 0.1306)
(7, 90, 0.1122)
(7, 100, 0.1029)

(8, 10, 0.8477)
(8, 20, 0.8042)
(8, 30, 0.7039)
(8, 40, 0.6008)
(8, 50, 0.4599)
(8, 60, 0.2615)
(8, 70, 0.1722)
(8, 80, 0.1331)
(8, 90, 0.1108)
(8, 100, 0.0947)

(9, 10, 0.8359)
(9, 20, 0.7932)
(9, 30, 0.7132)
(9, 40, 0.5425)
(9, 50, 0.3975)
(9, 60, 0.2580)
(9, 70, 0.1602)
(9, 80, 0.1257)
(9, 90, 0.0952)
(9, 100, 0.0283)

(10, 10, 0.8509)
(10, 20, 0.7797)
(10, 30, 0.7096)
(10, 40, 0.6056)
(10, 50, 0.4235)
(10, 60, 0.2568)
(10, 70, 0.1532)
(10, 80, 0.1080)
(10, 90, 0.0310)
(10, 100, 0.0131)

};

  \end{axis}
  \end{tikzpicture}

%% file: figures/3d_10_XGB.tex
\begin{tikzpicture}[xscale=1.0,yscale=1.0]
\begin{axis}[%
view={10}{26},
width=10cm,
height=6cm,
scale only axis,
xmin=1, xmax=10,
xmajorgrids,
ymin=10, ymax=100,
ymajorgrids,
zmin=0, zmax=1,
zmajorgrids,
axis lines=left,
grid=major,
xlabel=Adversarial clients,
xlabel style={yshift=0.2cm},
ylabel=Percent labels flipped,
ylabel style={yshift=0.5cm,xshift=0.15cm,rotate=55},
xtick={10, 9, 8, 7, 6, 5, 4, 3, 2, 1},
ytick={100, 70, 40, 10},
x dir=reverse,
y dir=reverse,
zlabel=Accuracy,
zlabel style={xshift=0.65cm,yshift=0.0cm},
x tick label style={xshift=0.05cm,scale=0.9,
    		/pgf/number format/.cd,
		1000 sep={},
    		fixed,
    		fixed zerofill,
    		precision=0},
y tick label style={yshift=0.25cm,xshift=0.075cm,scale=0.9,
    		/pgf/number format/.cd,
		1000 sep={},
    		fixed,
    		fixed zerofill,
    		precision=0},
z tick label style={yshift=0.4cm,scale=0.9,
    		/pgf/number format/.cd,
		1000 sep={},
    		fixed,
    		fixed zerofill,
    		precision=2}]
%
\addplot3[%
surf,
z buffer=sort,
colormap/jet,
shader=flat,
draw=black]
coordinates{ 

(10, 100, 0.0175)
(10, 90, 0.0431)
(10, 80, 0.1008)
(10, 70, 0.176)
(10, 60, 0.2308)
(10, 50, 0.3909)
(10, 40, 0.6002)
(10, 30, 0.6766)
(10, 20, 0.751)
(10, 10, 0.7954)

(9, 100, 0.0242)
(9, 90, 0.0786)
(9, 80, 0.1328)
(9, 70, 0.173)
(9, 60, 0.2527)
(9, 50, 0.4657)
(9, 40, 0.6013)
(9, 30, 0.679)
(9, 20, 0.7519)
(9, 10, 0.8148)

(8, 100, 0.0858)
(8, 90, 0.1181)
(8, 80, 0.1422)
(8, 70, 0.1946)
(8, 60, 0.2738)
(8, 50, 0.4065)
(8, 40, 0.6151)
(8, 30, 0.6832)
(8, 20, 0.7399)
(8, 10, 0.7978)

(7, 100, 0.1028)
(7, 90, 0.1233)
(7, 80, 0.1478)
(7, 70, 0.1801)
(7, 60, 0.2498)
(7, 50, 0.445)
(7, 40, 0.5973)
(7, 30, 0.6822)
(7, 20, 0.7949)
(7, 10, 0.8123)

(6, 100, 0.1375)
(6, 90, 0.1398)
(6, 80, 0.1492)
(6, 70, 0.1876)
(6, 60, 0.2795)
(6, 50, 0.4552)
(6, 40, 0.5988)
(6, 30, 0.702)
(6, 20, 0.8025)
(6, 10, 0.7988)

(5, 100, 0.1289)
(5, 90, 0.1367)
(5, 80, 0.1497)
(5, 70, 0.3228)
(5, 60, 0.5639)
(5, 50, 0.6939)
(5, 40, 0.7619)
(5, 30, 0.7705)
(5, 20, 0.7747)
(5, 10, 0.8105)

(4, 100, 0.3225)
(4, 90, 0.5282)
(4, 80, 0.6495)
(4, 70, 0.7587)
(4, 60, 0.7603)
(4, 50, 0.7867)
(4, 40, 0.7854)
(4, 30, 0.7908)
(4, 20, 0.8098)
(4, 10, 0.8161)

(3, 100, 0.7525)
(3, 90, 0.7557)
(3, 80, 0.7613)
(3, 70, 0.7624)
(3, 60, 0.7749)
(3, 50, 0.7792)
(3, 40, 0.7835)
(3, 30, 0.8072)
(3, 20, 0.8127)
(3, 10, 0.8172)

(2, 100, 0.7635)
(2, 90, 0.7714)
(2, 80, 0.7746)
(2, 70, 0.7868)
(2, 60, 0.7884)
(2, 50, 0.8068)
(2, 40, 0.8075)
(2, 30, 0.8143)
(2, 20, 0.8153)
(2, 10, 0.8176)

(1, 100, 0.8155)
(1, 90, 0.8147)
(1, 80, 0.82)
(1, 70, 0.8228)
(1, 60, 0.8161)
(1, 50, 0.8302)
(1, 40, 0.8168)
(1, 30, 0.8314)
(1, 20, 0.8282)
(1, 10, 0.832)

};
  \end{axis}
  \end{tikzpicture}

%% file: figures/3d_10_LSTM.tex
\begin{tikzpicture}[xscale=1.0,yscale=1.0]
\begin{axis}[%
view={10}{26},
width=10cm,
height=6cm,
scale only axis,
xmin=1, xmax=10,
xmajorgrids,
ymin=10, ymax=100,
ymajorgrids,
zmin=0.0, zmax=1.0,
zmajorgrids,
axis lines=left,
grid=major,
xlabel=Adversarial clients,
xlabel style={yshift=0.2cm},
ylabel=Percent labels flipped,
ylabel style={yshift=0.5cm,xshift=0.15cm,rotate=55},
xtick={10, 9, 8, 7, 6, 5, 4, 3, 2, 1},
ytick={100, 70, 40, 10},
x dir=reverse,
y dir=reverse,
zlabel=Accuracy,
zlabel style={xshift=0.65cm,yshift=0.0cm},
x tick label style={xshift=0.05cm,scale=0.9,
    		/pgf/number format/.cd,
		1000 sep={},
    		fixed,
    		fixed zerofill,
    		precision=0},
y tick label style={yshift=0.25cm,xshift=0.075cm,scale=0.9,
    		/pgf/number format/.cd,
		1000 sep={},
    		fixed,
    		fixed zerofill,
    		precision=0},
z tick label style={yshift=0.4cm,scale=0.9,
    		/pgf/number format/.cd,
		1000 sep={},
    		fixed,
    		fixed zerofill,
    		precision=2}]
%
\addplot3[%
surf,
z buffer=sort,
colormap/jet,
shader=flat,
draw=black]
coordinates{ 

(10, 100, 0.002)
(10, 90, 0.0017)
(10, 80, 0.0017)
(10, 70, 0.0019)
(10, 60, 0.0016)
(10, 50, 0.0018)
(10, 40, 0.9819)
(10, 30, 0.9856)
(10, 20, 0.9873)
(10, 10, 0.9869)

(9, 100, 0.0026)
(9, 90, 0.0029)
(9, 80, 0.0021)
(9, 70, 0.0022)
(9, 60, 0.0036)
(9, 50, 0.0024)
(9, 40, 0.9842)
(9, 30, 0.9855)
(9, 20, 0.9857)
(9, 10, 0.9867)

(8, 100, 0.0038)
(8, 90, 0.0049)
(8, 80, 0.0047)
(8, 70, 0.0051)
(8, 60, 0.0036)
(8, 50, 0.0046)
(8, 40, 0.9852)
(8, 30, 0.9858)
(8, 20, 0.9869)
(8, 10, 0.9877)

(7, 100, 0.0256)
(7, 90, 0.0401)
(7, 80, 0.0151)
(7, 70, 0.1097)
(7, 60, 0.1035)
(7, 50, 0.0405)
(7, 40, 0.9846)
(7, 30, 0.9866)
(7, 20, 0.9872)
(7, 10, 0.9868)

(6, 100, 0.1847)
(6, 90, 0.259)
(6, 80, 0.1627)
(6, 70, 0.1198)
(6, 60, 0.3325)
(6, 50, 0.133)
(6, 40, 0.9867)
(6, 30, 0.9872)
(6, 20, 0.985)
(6, 10, 0.987)

(5, 100, 0.6136)
(5, 90, 0.6287)
(5, 80, 0.4968)
(5, 70, 0.653)
(5, 60, 0.6435)
(5, 50, 0.6622)
(5, 40, 0.9853)
(5, 30, 0.9869)
(5, 20, 0.9869)
(5, 10, 0.9864)

(4, 100, 0.8803)
(4, 90, 0.858)
(4, 80, 0.8989)
(4, 70, 0.9117)
(4, 60, 0.9069)
(4, 50, 0.9405)
(4, 40, 0.9873)
(4, 30, 0.9872)
(4, 20, 0.9867)
(4, 10, 0.9873)

(3, 100, 0.9784)
(3, 90, 0.9792)
(3, 80, 0.9802)
(3, 70, 0.9797)
(3, 60, 0.9777)
(3, 50, 0.9775)
(3, 40, 0.9862)
(3, 30, 0.9877)
(3, 20, 0.9866)
(3, 10, 0.9871)

(2, 100, 0.9841)
(2, 90, 0.9845)
(2, 80, 0.9854)
(2, 70, 0.9838)
(2, 60, 0.9853)
(2, 50, 0.9844)
(2, 40, 0.987)
(2, 30, 0.986)
(2, 20, 0.9865)
(2, 10, 0.9874)

(1, 100, 0.9855)
(1, 90, 0.9859)
(1, 80, 0.9869)
(1, 70, 0.9859)
(1, 60, 0.9862)
(1, 50, 0.986)
(1, 40, 0.9866)
(1, 30, 0.9866)
(1, 20, 0.9871)
(1, 10, 0.9865)

};
  \end{axis}
  \end{tikzpicture}

%% file: figures/3d_100_MLR.tex
\begin{tikzpicture}[xscale=1.0,yscale=1.0]
\begin{axis}[%
view={10}{26},
width=10cm,
height=6cm,
scale only axis,
xmin=10, xmax=100,
xmajorgrids,
ymin=10, ymax=100,
ymajorgrids,
zmin=0.0, zmax=1.0,
zmajorgrids,
axis lines=left,
grid=major,
xlabel=Adversarial clients,
xlabel style={yshift=0.2cm},
ylabel=Percent labels flipped,
ylabel style={yshift=0.5cm,xshift=0.15cm,rotate=55},
xtick={100, 90, 80, 70, 60, 50, 40, 30, 20, 10},
ytick={100, 70, 40, 10},
x dir=reverse,
y dir=reverse,
zlabel=Accuracy,
zlabel style={xshift=0.65cm,yshift=0.0cm},
x tick label style={xshift=0.05cm,scale=0.9,
    		/pgf/number format/.cd,
		1000 sep={},
    		fixed,
    		fixed zerofill,
    		precision=0},
y tick label style={yshift=0.25cm,xshift=0.075cm,scale=0.9,
    		/pgf/number format/.cd,
		1000 sep={},
    		fixed,
    		fixed zerofill,
    		precision=0},
z tick label style={yshift=0.4cm,scale=0.9,
    		/pgf/number format/.cd,
		1000 sep={},
    		fixed,
    		fixed zerofill,
    		precision=2}]
%
\addplot3[%
surf,
z buffer=sort,
colormap/jet,
shader=flat,
draw=black]
coordinates{ 

(100, 100, 0.0097)
(100, 90, 0.019)
(100, 80, 0.0312)
(100, 70, 0.083)
(100, 60, 0.1962)
(100, 50, 0.3986)
(100, 40, 0.6643)
(100, 30, 0.7712)
(100, 20, 0.8373)
(100, 10, 0.8578)

(90, 100, 0.032)
(90, 90, 0.1667)
(90, 80, 0.2296)
(90, 70, 0.2717)
(90, 60, 0.4698)
(90, 50, 0.6524)
(90, 40, 0.7147)
(90, 30, 0.8119)
(90, 20, 0.849)
(90, 10, 0.8644)

(80, 100, 0.2624)
(80, 90, 0.332)
(80, 80, 0.3764)
(80, 70, 0.448)
(80, 60, 0.5581)
(80, 50, 0.712)
(80, 40, 0.7492)
(80, 30, 0.8287)
(80, 20, 0.8601)
(80, 10, 0.8637)

(70, 100, 0.2312)
(70, 90, 0.3429)
(70, 80, 0.4059)
(70, 70, 0.5635)
(70, 60, 0.631)
(70, 50, 0.7331)
(70, 40, 0.7945)
(70, 30, 0.8355)
(70, 20, 0.8482)
(70, 10, 0.8707)

(60, 100, 0.2792)
(60, 90, 0.4523)
(60, 80, 0.5433)
(60, 70, 0.6017)
(60, 60, 0.6999)
(60, 50, 0.7738)
(60, 40, 0.7943)
(60, 30, 0.8253)
(60, 20, 0.831)
(60, 10, 0.8707)

(50, 100, 0.3666)
(50, 90, 0.5579)
(50, 80, 0.6003)
(50, 70, 0.6918)
(50, 60, 0.7625)
(50, 50, 0.7815)
(50, 40, 0.7994)
(50, 30, 0.8462)
(50, 20, 0.8578)
(50, 10, 0.862)

(40, 100, 0.469)
(40, 90, 0.6029)
(40, 80, 0.6834)
(40, 70, 0.7373)
(40, 60, 0.7724)
(40, 50, 0.841)
(40, 40, 0.8376)
(40, 30, 0.8399)
(40, 20, 0.8651)
(40, 10, 0.8658)

(30, 100, 0.552)
(30, 90, 0.6976)
(30, 80, 0.7513)
(30, 70, 0.7958)
(30, 60, 0.7955)
(30, 50, 0.8331)
(30, 40, 0.8304)
(30, 30, 0.8755)
(30, 20, 0.8673)
(30, 10, 0.8769)

(20, 100, 0.6003)
(20, 90, 0.7425)
(20, 80, 0.7815)
(20, 70, 0.8171)
(20, 60, 0.8232)
(20, 50, 0.8324)
(20, 40, 0.8509)
(20, 30, 0.8739)
(20, 20, 0.8604)
(20, 10, 0.8712)

(10, 100, 0.7716)
(10, 90, 0.7736)
(10, 80, 0.7779)
(10, 70, 0.7943)
(10, 60, 0.8518)
(10, 50, 0.8497)
(10, 40, 0.8742)
(10, 30, 0.8692)
(10, 20, 0.8766)
(10, 10, 0.8622)

};
  \end{axis}
  \end{tikzpicture}

%% file: figures/3d_100_SVC.tex
\begin{tikzpicture}[xscale=1.0,yscale=1.0]
\begin{axis}[%
view={10}{26},
width=10cm,
height=6cm,
scale only axis,
xmin=10, xmax=100,
xmajorgrids,
ymin=10, ymax=100,
ymajorgrids,
zmin=0.0, zmax=1.0,
zmajorgrids,
axis lines=left,
grid=major,
xlabel=Adversarial clients,
xlabel style={yshift=0.2cm},
ylabel=Percent labels flipped,
ylabel style={yshift=0.5cm,xshift=0.15cm,rotate=55},
xtick={100, 90, 80, 70, 60, 50, 40, 30, 20, 10},
ytick={100, 70, 40, 10},
x dir=reverse,
y dir=reverse,
zlabel=Accuracy,
zlabel style={xshift=0.65cm,yshift=0.0cm},
x tick label style={xshift=0.05cm,scale=0.9,
    		/pgf/number format/.cd,
		1000 sep={},
    		fixed,
    		fixed zerofill,
    		precision=0},
y tick label style={yshift=0.25cm,xshift=0.075cm,scale=0.9,
    		/pgf/number format/.cd,
		1000 sep={},
    		fixed,
    		fixed zerofill,
    		precision=0},
z tick label style={yshift=0.4cm,scale=0.9,
    		/pgf/number format/.cd,
		1000 sep={},
    		fixed,
    		fixed zerofill,
    		precision=2}]


\addplot3[%
surf,
z buffer=sort,
colormap/jet,
shader=flat,
draw=black]
coordinates{ 

(100, 100, 0.0219)
(100, 90, 0.0251)
(100, 80, 0.033)
(100, 70, 0.0527)
(100, 60, 0.1427)
(100, 50, 0.3973)
(100, 40, 0.6179)
(100, 30, 0.7507)
(100, 20, 0.7827)
(100, 10, 0.7769)

(90, 100, 0.0278)
(90, 90, 0.0313)
(90, 80, 0.0678)
(90, 70, 0.091)
(90, 60, 0.2463)
(90, 50, 0.4695)
(90, 40, 0.6345)
(90, 30, 0.7456)
(90, 20, 0.7778)
(90, 10, 0.7747)

(80, 100, 0.0329)
(80, 90, 0.0468)
(80, 80, 0.0856)
(80, 70, 0.2032)
(80, 60, 0.3848)
(80, 50, 0.6059)
(80, 40, 0.6985)
(80, 30, 0.7548)
(80, 20, 0.7791)
(80, 10, 0.7905)

(70, 100, 0.049)
(70, 90, 0.1187)
(70, 80, 0.2039)
(70, 70, 0.3845)
(70, 60, 0.5952)
(70, 50, 0.6874)
(70, 40, 0.7464)
(70, 30, 0.7456)
(70, 20, 0.7756)
(70, 10, 0.7685)

(60, 100, 0.136)
(60, 90, 0.312)
(60, 80, 0.36)
(60, 70, 0.5804)
(60, 60, 0.634)
(60, 50, 0.7038)
(60, 40, 0.7572)
(60, 30, 0.7724)
(60, 20, 0.7775)
(60, 10, 0.7827)

(50, 100, 0.3797)
(50, 90, 0.5387)
(50, 80, 0.684)
(50, 70, 0.7074)
(50, 60, 0.7293)
(50, 50, 0.7698)
(50, 40, 0.7781)
(50, 30, 0.7859)
(50, 20, 0.7691)
(50, 10, 0.7838)

(40, 100, 0.6981)
(40, 90, 0.7124)
(40, 80, 0.6749)
(40, 70, 0.72)
(40, 60, 0.6694)
(40, 50, 0.7562)
(40, 40, 0.7888)
(40, 30, 0.778)
(40, 20, 0.7767)
(40, 10, 0.7783)

(30, 100, 0.7475)
(30, 90, 0.7658)
(30, 80, 0.7547)
(30, 70, 0.762)
(30, 60, 0.7789)
(30, 50, 0.7896)
(30, 40, 0.7909)
(30, 30, 0.7958)
(30, 20, 0.7838)
(30, 10, 0.7626)

(20, 100, 0.7652)
(20, 90, 0.7852)
(20, 80, 0.7981)
(20, 70, 0.7872)
(20, 60, 0.7912)
(20, 50, 0.7918)
(20, 40, 0.7706)
(20, 30, 0.7896)
(20, 20, 0.7928)
(20, 10, 0.7918)

(10, 100, 0.7875)
(10, 90, 0.7878)
(10, 80, 0.7885)
(10, 70, 0.7878)
(10, 60, 0.7892)
(10, 50, 0.7786)
(10, 40, 0.7854)
(10, 30, 0.7916)
(10, 20, 0.7925)
(10, 10, 0.7747)

};
  \end{axis}
  \end{tikzpicture}

%% file: figures/3d_100_MLP.tex
\begin{tikzpicture}[xscale=1.0,yscale=1.0]
\begin{axis}[%
view={10}{26},
width=10cm,
height=6cm,
scale only axis,
xmin=10, xmax=100,
xmajorgrids,
ymin=10, ymax=100,
ymajorgrids,
zmin=0.0, zmax=1.0,
zmajorgrids,
axis lines=left,
grid=major,
xlabel=Adversarial clients,
xlabel style={yshift=0.2cm},
ylabel=Percent labels flipped,
ylabel style={yshift=0.5cm,xshift=0.15cm,rotate=55},
xtick={100, 90, 80, 70, 60, 50, 40, 30, 20, 10},
ytick={100, 70, 40, 10},
x dir=reverse,
y dir=reverse,
zlabel=Accuracy,
zlabel style={xshift=0.65cm,yshift=0.0cm},
x tick label style={xshift=0.05cm,scale=0.9,
    		/pgf/number format/.cd,
		1000 sep={},
    		fixed,
    		fixed zerofill,
    		precision=0},
y tick label style={yshift=0.25cm,xshift=0.075cm,scale=0.9,
    		/pgf/number format/.cd,
		1000 sep={},
    		fixed,
    		fixed zerofill,
    		precision=0},
z tick label style={yshift=0.4cm,scale=0.9,
    		/pgf/number format/.cd,
		1000 sep={},
    		fixed,
    		fixed zerofill,
    		precision=2}]
%
\addplot3[%
surf,
z buffer=sort,
colormap/jet,
shader=flat,
draw=black]
coordinates{ 

(100, 100, 0.0071)
(100, 90, 0.0047)
(100, 80, 0.0061)
(100, 70, 0.0065)
(100, 60, 0.0049)
(100, 50, 0.0063)
(100, 40, 0.0103)
(100, 30, 0.0421)
(100, 20, 0.4358)
(100, 10, 0.8922)

(90, 100, 0.0067)
(90, 90, 0.0076)
(90, 80, 0.0067)
(90, 70, 0.0085)
(90, 60, 0.0053)
(90, 50, 0.006)
(90, 40, 0.0179)
(90, 30, 0.0558)
(90, 20, 0.6851)
(90, 10, 0.8979)

(80, 100, 0.0147)
(80, 90, 0.014)
(80, 80, 0.0167)
(80, 70, 0.0187)
(80, 60, 0.0133)
(80, 50, 0.0177)
(80, 40, 0.0343)
(80, 30, 0.2688)
(80, 20, 0.8356)
(80, 10, 0.9004)

(70, 100, 0.0346)
(70, 90, 0.0327)
(70, 80, 0.0309)
(70, 70, 0.0361)
(70, 60, 0.0409)
(70, 50, 0.049)
(70, 40, 0.1339)
(70, 30, 0.5336)
(70, 20, 0.8512)
(70, 10, 0.9009)

(60, 100, 0.2384)
(60, 90, 0.254)
(60, 80, 0.1015)
(60, 70, 0.1325)
(60, 60, 0.148)
(60, 50, 0.0813)
(60, 40, 0.2659)
(60, 30, 0.8083)
(60, 20, 0.8854)
(60, 10, 0.904)

(50, 100, 0.6046)
(50, 90, 0.5705)
(50, 80, 0.5955)
(50, 70, 0.4885)
(50, 60, 0.3949)
(50, 50, 0.4621)
(50, 40, 0.7589)
(50, 30, 0.8755)
(50, 20, 0.8921)
(50, 10, 0.9069)

(40, 100, 0.8095)
(40, 90, 0.7777)
(40, 80, 0.7735)
(40, 70, 0.7588)
(40, 60, 0.848)
(40, 50, 0.8428)
(40, 40, 0.8589)
(40, 30, 0.912)
(40, 20, 0.9084)
(40, 10, 0.9075)

(30, 100, 0.8976)
(30, 90, 0.8988)
(30, 80, 0.9014)
(30, 70, 0.8904)
(30, 60, 0.9013)
(30, 50, 0.903)
(30, 40, 0.9106)
(30, 30, 0.9105)
(30, 20, 0.9185)
(30, 10, 0.919)

(20, 100, 0.9158)
(20, 90, 0.9164)
(20, 80, 0.9285)
(20, 70, 0.9298)
(20, 60, 0.9271)
(20, 50, 0.9113)
(20, 40, 0.9255)
(20, 30, 0.9272)
(20, 20, 0.9294)
(20, 10, 0.9214)

(10, 100, 0.9325)
(10, 90, 0.9337)
(10, 80, 0.9303)
(10, 70, 0.9329)
(10, 60, 0.9335)
(10, 50, 0.9384)
(10, 40, 0.934)
(10, 30, 0.9388)
(10, 20, 0.9373)
(10, 10, 0.9351)

};
  \end{axis}
  \end{tikzpicture}

%% file: figures/3d_100_CNN.tex
\begin{tikzpicture}[xscale=1.0,yscale=1.0]
\begin{axis}[%
view={10}{26},
width=10cm,
height=6cm,
scale only axis,
xmin=10, xmax=100,
xmajorgrids,
ymin=10, ymax=100,
ymajorgrids,
zmin=0.0, zmax=1.0,
zmajorgrids,
axis lines=left,
grid=major,
xlabel=Adversarial clients,
xlabel style={yshift=0.2cm},
ylabel=Percent labels flipped,
ylabel style={yshift=0.5cm,xshift=0.15cm,rotate=55},
xtick={100, 90, 80, 70, 60, 50, 40, 30, 20, 10},
ytick={100, 70, 40, 10},
x dir=reverse,
y dir=reverse,
zlabel=Accuracy,
zlabel style={xshift=0.65cm,yshift=0.0cm},
x tick label style={xshift=0.05cm,scale=0.9,
    		/pgf/number format/.cd,
		1000 sep={},
    		fixed,
    		fixed zerofill,
    		precision=0},
y tick label style={yshift=0.25cm,xshift=0.075cm,scale=0.9,
    		/pgf/number format/.cd,
		1000 sep={},
    		fixed,
    		fixed zerofill,
    		precision=0},
z tick label style={yshift=0.4cm,scale=0.9,
    		/pgf/number format/.cd,
		1000 sep={},
    		fixed,
    		fixed zerofill,
    		precision=2}]
%
\addplot3[%
surf,
z buffer=sort,
colormap/jet,
shader=flat,
draw=black]
coordinates{ 

(100, 100, 0.1348)
(100, 90, 0.0995)
(100, 80, 0.0584)
(100, 70, 0.1033)
(100, 60, 0.0775)
(100, 50, 0.1030)
(100, 40, 0.2190)
(100, 30, 0.1368)
(100, 20, 0.3337)
(100, 10, 0.3099)

(90, 100, 0.1081)
(90, 90, 0.0475)
(90, 80, 0.0389)
(90, 70, 0.0795)
(90, 60, 0.0832)
(90, 50, 0.0937)
(90, 40, 0.2093)
(90, 30, 0.3987)
(90, 20, 0.3642)
(90, 10, 0.6478)

(80, 100, 0.1180)
(80, 90, 0.1125)
(80, 80, 0.0901)
(80, 70, 0.1486)
(80, 60, 0.0608)
(80, 50, 0.0790)
(80, 40, 0.3309)
(80, 30, 0.3951)
(80, 20, 0.5991)
(80, 10, 0.8391)

(70, 100, 0.2234)
(70, 90, 0.2300)
(70, 80, 0.1516)
(70, 70, 0.2414)
(70, 60, 0.1810)
(70, 50, 0.1963)
(70, 40, 0.4537)
(70, 30, 0.4980)
(70, 20, 0.7251)
(70, 10, 0.8748)

(60, 100, 0.3636)
(60, 90, 0.3665)
(60, 80, 0.3192)
(60, 70, 0.4458)
(60, 60, 0.3323)
(60, 50, 0.3463)
(60, 40, 0.5939)
(60, 30, 0.7320)
(60, 20, 0.8671)
(60, 10, 0.8875)

(50, 100, 0.5534)
(50, 90, 0.4215)
(50, 80, 0.4558)
(50, 70, 0.5027)
(50, 60, 0.5062)
(50, 50, 0.4785)
(50, 40, 0.7271)
(50, 30, 0.7474)
(50, 20, 0.7457)
(50, 10, 0.7336)

(40, 100, 0.3712)
(40, 90, 0.3382)
(40, 80, 0.3404)
(40, 70, 0.4578)
(40, 60, 0.2380)
(40, 50, 0.3864)
(40, 40, 0.5851)
(40, 30, 0.5489)
(40, 20, 0.5158)
(40, 10, 0.5225)

(30, 100, 0.4783)
(30, 90, 0.5107)
(30, 80, 0.5169)
(30, 70, 0.5763)
(30, 60, 0.5535)
(30, 50, 0.5569)
(30, 40, 0.7451)
(30, 30, 0.7205)
(30, 20, 0.6916)
(30, 10, 0.7446)

(20, 100, 0.8815)
(20, 90, 0.8928)
(20, 80, 0.9007)
(20, 70, 0.8650)
(20, 60, 0.8820)
(20, 50, 0.8434)
(20, 40, 0.8311)
(20, 30, 0.9165)
(20, 20, 0.9211)
(20, 10, 0.8911)

(10, 100, 0.9315)
(10, 90, 0.9309)
(10, 80, 0.9337)
(10, 70, 0.9398)
(10, 60, 0.9276)
(10, 50, 0.9232)
(10, 40, 0.9341)
(10, 30, 0.9309)
(10, 20, 0.9269)
(10, 10, 0.9360)

};
  \end{axis}
  \end{tikzpicture}

%% file: figures/3d_100_RF.tex
\begin{tikzpicture}[xscale=1.0,yscale=1.0]
\begin{axis}[%
view={10}{26},
width=10cm,
height=6cm,
scale only axis,
xmin=10, xmax=100,
xmajorgrids,
ymin=10, ymax=100,
ymajorgrids,
zmin=0.1, zmax=1,
zmajorgrids,
axis lines=left,
grid=major,
xlabel=Adversarial clients,
xlabel style={yshift=0.2cm},
ylabel=Percent labels flipped,
ylabel style={yshift=0.5cm,xshift=0.15cm,rotate=55},
xtick={100, 90, 80, 70, 60, 50, 40, 30, 20, 10},
ytick={100, 70, 40, 10},
x dir=reverse,
y dir=reverse,
zlabel=Accuracy,
zlabel style={xshift=0.65cm,yshift=0.0cm},
x tick label style={xshift=0.05cm,scale=0.9,
    		/pgf/number format/.cd,
		1000 sep={},
    		fixed,
    		fixed zerofill,
    		precision=0},
y tick label style={yshift=0.25cm,xshift=0.075cm,scale=0.9,
    		/pgf/number format/.cd,
		1000 sep={},
    		fixed,
    		fixed zerofill,
    		precision=0},
z tick label style={yshift=0.4cm,scale=0.9,
    		/pgf/number format/.cd,
		1000 sep={},
    		fixed,
    		fixed zerofill,
    		precision=2}]
%
\addplot3[%
surf,
z buffer=sort,
colormap/jet,
shader=flat,
draw=black]
coordinates{ 

(10, 10, 0.7011)
(10, 20, 0.6739)
(10, 30, 0.7226)
(10, 40, 0.7063)
(10, 50, 0.7101)
(10, 60, 0.6783)
(10, 70, 0.6631)
(10, 80, 0.7220)
(10, 90, 0.7164)
(10, 100, 0.6800)

(20, 10, 0.6861)
(20, 20, 0.6701)
(20, 30, 0.6841)
(20, 40, 0.6851)
(20, 50, 0.7036)
(20, 60, 0.6891)
(20, 70, 0.68595)
(20, 80, 0.6634)
(20, 90, 0.6500)
(20, 100, 0.6706)

(30, 10, 0.6882)
(30, 20, 0.7077)
(30, 30, 0.6877)
(30, 40, 0.7161)
(30, 50, 0.6818)
(30, 60, 0.6496)
(30, 70, 0.6511)
(30, 80, 0.6302)
(30, 90, 0.6217)
(30, 100, 0.6185)

(40, 10, 0.7080)
(40, 20, 0.6777)
(40, 30, 0.6966)
(40, 40, 0.6624)
(40, 50, 0.6670)
(40, 60, 0.6472)
(40, 70, 0.6573)
(40, 80, 0.6635)
(40, 90, 0.1353)
(40, 100, 0.1198)

(50, 10, 0.7075)
(50, 20, 0.6524)
(50, 30, 0.631)
(50, 40, 0.6713)
(50, 50, 0.6584)
(50, 60, 0.2947)
(50, 70, 0.2634)
(50, 80, 0.1623)
(50, 90, 0.1202)
(50, 100, 0.1167)

(60, 10, 0.6884)
(60, 20, 0.7300)
(60, 30, 0.6919)
(60, 40, 0.6882)
(60, 50, 0.4015)
(60, 60, 0.2790)
(60, 70, 0.2312)
(60, 80, 0.1392)
(60, 90, 0.1323)
(60, 100, 0.1158)

(70, 10, 0.7313)
(70, 20, 0.6321)
(70, 30, 0.6191)
(70, 40, 0.4923)
(70, 50, 0.3993)
(70, 60, 0.2806)
(70, 70, 0.2039)
(70, 80, 0.1327)
(70, 90, 0.1234)
(70, 100, 0.1158)

(80, 10, 0.6946)
(80, 20, 0.6308)
(80, 30, 0.5378)
(80, 40, 0.4447)
(80, 50, 0.3826)
(80, 60, 0.3201)
(80, 70, 0.2003)
(80, 80, 0.1430)
(80, 90, 0.1208)
(80, 100, 0.1074)

(90, 10, 0.6941)
(90, 20, 0.6520)
(90, 30, 0.5904)
(90, 40, 0.4532)
(90, 50, 0.3775)
(90, 60, 0.2851)
(90, 70, 0.1793)
(90, 80, 0.1648)
(90, 90, 0.1102)
(90, 100, 0.0975)

(100, 10, 0.6934)
(100, 20, 0.6789)
(100, 30, 0.5948)
(100, 40, 0.4826)
(100, 50, 0.3546)
(100, 60, 0.2461)
(100, 70, 0.1592)
(100, 80, 0.1046)
(100, 90, 0.0415)
(100, 100, 0.0234)

};

  \end{axis}
  \end{tikzpicture}

%% file: figures/3d_100_XGB.tex
\begin{tikzpicture}[xscale=1.0,yscale=1.0]
\begin{axis}[%
view={10}{26},
width=10cm,
height=6cm,
scale only axis,
xmin=10, xmax=100,
xmajorgrids,
ymin=10, ymax=100,
ymajorgrids,
zmin=0, zmax=1,
zmajorgrids,
axis lines=left,
grid=major,
xlabel=Adversarial clients,
xlabel style={yshift=0.2cm},
ylabel=Percent labels flipped,
ylabel style={yshift=0.5cm,xshift=0.15cm,rotate=55},
xtick={100, 90, 80, 70, 60, 50, 40, 30, 20, 10},
ytick={100, 70, 40, 10},
x dir=reverse,
y dir=reverse,
zlabel=Accuracy,
zlabel style={xshift=0.65cm,yshift=0.0cm},
x tick label style={xshift=0.05cm,scale=0.9,
    		/pgf/number format/.cd,
		1000 sep={},
    		fixed,
    		fixed zerofill,
    		precision=0},
y tick label style={yshift=0.25cm,xshift=0.075cm,scale=0.9,
    		/pgf/number format/.cd,
		1000 sep={},
    		fixed,
    		fixed zerofill,
    		precision=0},
z tick label style={yshift=0.4cm,scale=0.9,
    		/pgf/number format/.cd,
		1000 sep={},
    		fixed,
    		fixed zerofill,
    		precision=2}]
%
\addplot3[%
surf,
z buffer=sort,
colormap/jet,
shader=flat,
draw=black]
coordinates{ 

(100, 100, 0.0329)
(100, 90, 0.0738)
(100, 80, 0.1312)
(100, 70, 0.1841)
(100, 60, 0.2815)
(100, 50, 0.3498)
(100, 40, 0.3744)
(100, 30, 0.422)
(100, 20, 0.4556)
(100, 10, 0.5325)

(90, 100, 0.0537)
(90, 90, 0.0982)
(90, 80, 0.1557)
(90, 70, 0.2081)
(90, 60, 0.2827)
(90, 50, 0.3637)
(90, 40, 0.4864)
(90, 30, 0.5145)
(90, 20, 0.524)
(90, 10, 0.548)

(80, 100, 0.0904)
(80, 90, 0.1295)
(80, 80, 0.1768)
(80, 70, 0.2033)
(80, 60, 0.2845)
(80, 50, 0.3578)
(80, 40, 0.4196)
(80, 30, 0.5159)
(80, 20, 0.5233)
(80, 10, 0.593)

(70, 100, 0.1044)
(70, 90, 0.1482)
(70, 80, 0.257)
(70, 70, 0.2562)
(70, 60, 0.299)
(70, 50, 0.3977)
(70, 40, 0.4373)
(70, 30, 0.522)
(70, 20, 0.5962)
(70, 10, 0.6433)

(60, 100, 0.1924)
(60, 90, 0.1372)
(60, 80, 0.1841)
(60, 70, 0.2965)
(60, 60, 0.3176)
(60, 50, 0.4451)
(60, 40, 0.4505)
(60, 30, 0.6481)
(60, 20, 0.5594)
(60, 10, 0.6464)

(50, 100, 0.1228)
(50, 90, 0.2454)
(50, 80, 0.3178)
(50, 70, 0.3301)
(50, 60, 0.4257)
(50, 50, 0.5737)
(50, 40, 0.6409)
(50, 30, 0.5295)
(50, 20, 0.659)
(50, 10, 0.6269)

(40, 100, 0.3923)
(40, 90, 0.4507)
(40, 80, 0.5887)
(40, 70, 0.63)
(40, 60, 0.6375)
(40, 50, 0.6391)
(40, 40, 0.5645)
(40, 30, 0.5489)
(40, 20, 0.6467)
(40, 10, 0.6624)

(30, 100, 0.6261)
(30, 90, 0.6183)
(30, 80, 0.6332)
(30, 70, 0.6357)
(30, 60, 0.6296)
(30, 50, 0.668)
(30, 40, 0.6757)
(30, 30, 0.6539)
(30, 20, 0.6411)
(30, 10, 0.6823)

(20, 100, 0.6542)
(20, 90, 0.6254)
(20, 80, 0.6472)
(20, 70, 0.6357)
(20, 60, 0.6678)
(20, 50, 0.6422)
(20, 40, 0.6718)
(20, 30, 0.6384)
(20, 20, 0.6871)
(20, 10, 0.6524)

(10, 100, 0.6346)
(10, 90, 0.6518)
(10, 80, 0.6458)
(10, 70, 0.6568)
(10, 60, 0.6582)
(10, 50, 0.6477)
(10, 40, 0.6717)
(10, 30, 0.6852)
(10, 20, 0.6496)
(10, 10, 0.6623)

};
  \end{axis}
  \end{tikzpicture}

%% file: figures/3d_100_LSTM.tex
\begin{tikzpicture}[xscale=1.0,yscale=1.0]
\begin{axis}[%
view={10}{26},
width=10cm,
height=6cm,
scale only axis,
xmin=10, xmax=100,
xmajorgrids,
ymin=10, ymax=100,
ymajorgrids,
zmin=0.0, zmax=1.0,
zmajorgrids,
axis lines=left,
grid=major,
xlabel=Adversarial clients,
xlabel style={yshift=0.2cm},
ylabel=Percent labels flipped,
ylabel style={yshift=0.5cm,xshift=0.15cm,rotate=55},
xtick={100, 90, 80, 70, 60, 50, 40, 30, 20, 10},
ytick={100, 70, 40, 10},
x dir=reverse,
y dir=reverse,
zlabel=Accuracy,
zlabel style={xshift=0.65cm,yshift=0.0cm},
x tick label style={xshift=0.05cm,scale=0.9,
    		/pgf/number format/.cd,
		1000 sep={},
    		fixed,
    		fixed zerofill,
    		precision=0},
y tick label style={yshift=0.25cm,xshift=0.075cm,scale=0.9,
    		/pgf/number format/.cd,
		1000 sep={},
    		fixed,
    		fixed zerofill,
    		precision=0},
z tick label style={yshift=0.4cm,scale=0.9,
    		/pgf/number format/.cd,
		1000 sep={},
    		fixed,
    		fixed zerofill,
    		precision=2}]
%
\addplot3[%
surf,
z buffer=sort,
colormap/jet,
shader=flat,
draw=black]
coordinates{ 

(100, 100, 0.0049)
(100, 90, 0.0046)
(100, 80, 0.0045)
(100, 70, 0.0037)
(100, 60, 0.0052)
(100, 50, 0.0049)
(100, 40, 0.718)
(100, 30, 0.8978)
(100, 20, 0.933)
(100, 10, 0.9356)

(90, 100, 0.0065)
(90, 90, 0.0077)
(90, 80, 0.0087)
(90, 70, 0.0066)
(90, 60, 0.0061)
(90, 50, 0.0129)
(90, 40, 0.8423)
(90, 30, 0.918)
(90, 20, 0.9371)
(90, 10, 0.939)

(80, 100, 0.0133)
(80, 90, 0.0093)
(80, 80, 0.0154)
(80, 70, 0.0127)
(80, 60, 0.0163)
(80, 50, 0.052)
(80, 40, 0.7544)
(80, 30, 0.9248)
(80, 20, 0.9346)
(80, 10, 0.9415)

(70, 100, 0.0501)
(70, 90, 0.029)
(70, 80, 0.0407)
(70, 70, 0.0439)
(70, 60, 0.0625)
(70, 50, 0.1432)
(70, 40, 0.8833)
(70, 30, 0.9305)
(70, 20, 0.935)
(70, 10, 0.9436)

(60, 100, 0.1925)
(60, 90, 0.1464)
(60, 80, 0.2659)
(60, 70, 0.2192)
(60, 60, 0.1509)
(60, 50, 0.3682)
(60, 40, 0.9252)
(60, 30, 0.9283)
(60, 20, 0.9407)
(60, 10, 0.9433)

(50, 100, 0.4718)
(50, 90, 0.5198)
(50, 80, 0.4431)
(50, 70, 0.4406)
(50, 60, 0.458)
(50, 50, 0.5516)
(50, 40, 0.9377)
(50, 30, 0.9393)
(50, 20, 0.9413)
(50, 10, 0.945)

(40, 100, 0.8031)
(40, 90, 0.753)
(40, 80, 0.7254)
(40, 70, 0.6716)
(40, 60, 0.7954)
(40, 50, 0.8627)
(40, 40, 0.9251)
(40, 30, 0.9402)
(40, 20, 0.9453)
(40, 10, 0.946)

(30, 100, 0.8413)
(30, 90, 0.8701)
(30, 80, 0.8985)
(30, 70, 0.9122)
(30, 60, 0.8939)
(30, 50, 0.9166)
(30, 40, 0.9404)
(30, 30, 0.9424)
(30, 20, 0.9445)
(30, 10, 0.9478)

(20, 100, 0.9175)
(20, 90, 0.9357)
(20, 80, 0.9348)
(20, 70, 0.944)
(20, 60, 0.9368)
(20, 50, 0.9405)
(20, 40, 0.9464)
(20, 30, 0.9469)
(20, 20, 0.9471)
(20, 10, 0.9473)

(10, 100, 0.9433)
(10, 90, 0.9478)
(10, 80, 0.9428)
(10, 70, 0.9469)
(10, 60, 0.9428)
(10, 50, 0.9423)
(10, 40, 0.9486)
(10, 30, 0.9489)
(10, 20, 0.9488)
(10, 10, 0.9492)

};
  \end{axis}
  \end{tikzpicture}